\documentclass[final]{cvpr}

\usepackage{times}
\usepackage{epsfig}
\usepackage{graphicx}
\usepackage{amsmath}
\usepackage{amssymb}

\usepackage{booktabs}
\usepackage{multirow}
\usepackage{color, colortbl}
\usepackage{subcaption}
\usepackage[table]{xcolor}
\usepackage{pifont}
\usepackage[pagebackref=true,breaklinks=true,colorlinks,bookmarks=false]{hyperref}

\def\cvprPaperID{2308} 
\def\confYear{CVPR 2021}
\newcommand{\argmax}{\mathop{\mathrm{argmax}}}
\begin{document}

\title{Can audio-visual integration strengthen robustness \\under multimodal attacks?}

\author{Yapeng Tian \qquad Chenliang Xu\\
University of Rochester\\
{\tt\small \{yapengtian,chenliang.xu\}@rochester.edu}
}

\maketitle

\begin{abstract}
In this paper, we propose to make a systematic study on machines’ multisensory perception under attacks. We use the audio-visual event recognition task against multimodal adversarial attacks as a proxy to investigate the robustness of audio-visual learning.
We attack audio, visual, and both modalities to explore whether audio-visual integration still strengthens perception and how different fusion mechanisms affect the robustness of audio-visual models. For interpreting the multimodal interactions under attacks, we learn a weakly-supervised sound source visual localization model to localize sounding regions in videos.
To mitigate multimodal attacks, we propose an audio-visual defense approach based on an audio-visual dissimilarity constraint and external feature memory banks. Extensive experiments demonstrate that audio-visual models are susceptible to multimodal adversarial attacks; audio-visual integration could decrease the model robustness rather than strengthen under multimodal attacks; even a weakly-supervised sound source visual localization model can be successfully fooled; our defense method can improve the invulnerability of audio-visual networks without significantly sacrificing clean model performance. {{The source code and pre-trained models are released in \small\url{https://github.com/YapengTian/AV-Robustness-CVPR21}}}.    
\end{abstract}

\definecolor{Gray}{gray}{0.9}
\definecolor{LightCyan}{rgb}{0.88,1,1}

\section{Introduction}
\label{sec:intro}

\begin{figure}
    \centering
    \includegraphics[width=0.91\linewidth]{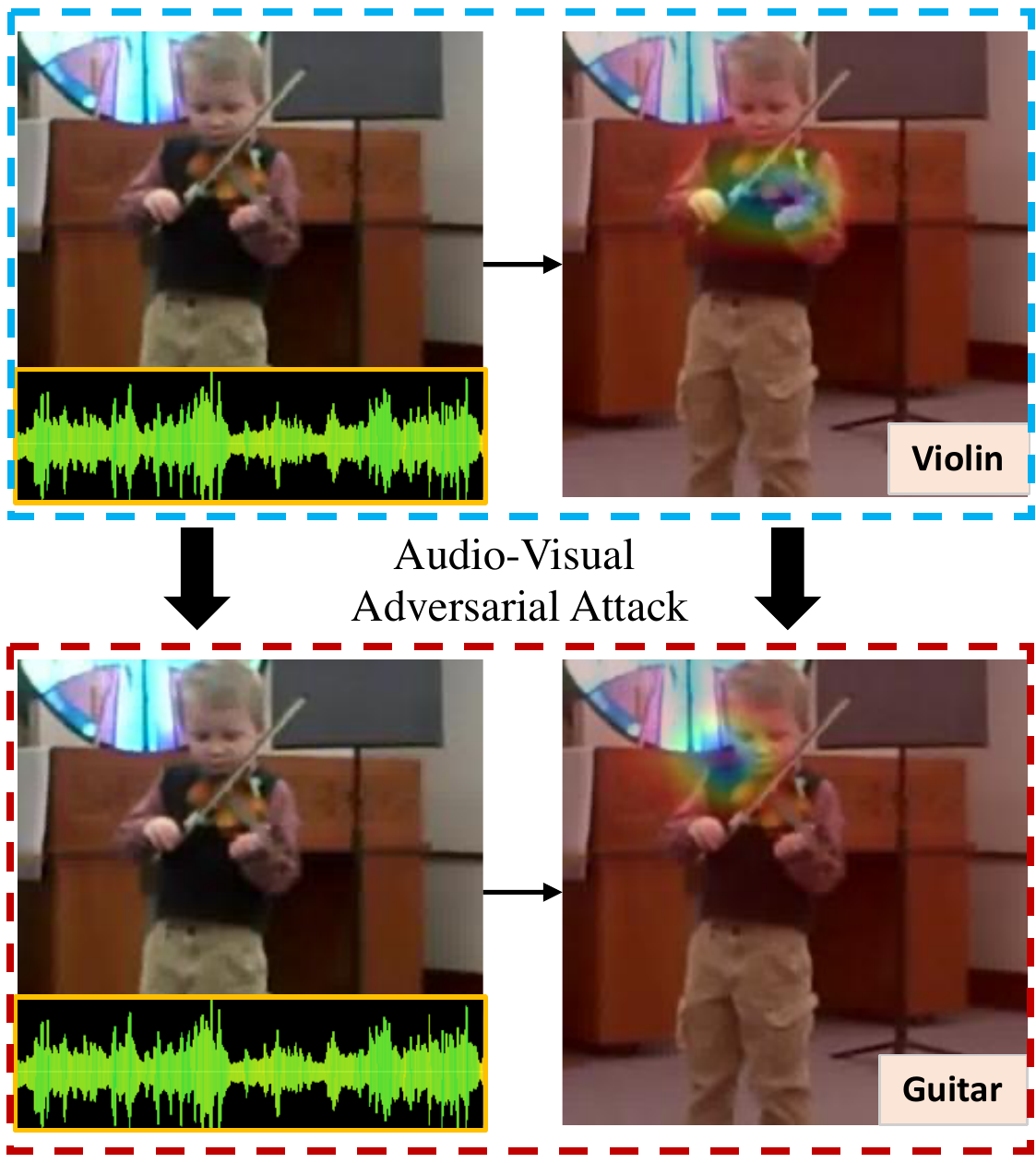}
    \vspace{-3mm}
    \caption{Adding imperceptible perturbations into audio and visual inputs by an audio-visual adversarial attack, our joint perception model predicts a wrong event class: \textit{Guitar} and tend to localize visual regions without the sound source. } 
    \label{fig:teaser}
    \vspace{-5mm}
\end{figure}

Our daily perceptual experiences are specified by multiple cooperated senses with multisensory integration~\cite{murray2011neural}. When we are talking with a person, we can learn her/his spoken words and emotions from the seen lip movements, gestures, facial expressions, and heard speech sounds. Numerous psychological and cognitive studies show that the availability of sensory inputs from several modalities ensures the robustness of the human perception system~\cite{sumby1954visual,gick2009aero,von2012multisensory}. However, the robustness highly depends on the reliability of multisensory inputs. For our human perception system, it might fail if certain senses are attacked. For example, the McGurk effect\footnote{\href{https://www.youtube.com/watch?v=2k8fHR9jKVM}{https://www.youtube.com/watch?v=2k8fHR9jKVM}}~\cite{mcgurk1976hearing} indicates a perceptual illusion, which occurs when a speech sound is paired with the visual component of another sound, leading to the perception of a third speech sound.

For computation models, our community indeed has devoted to develop data-driven approaches in lip reading~\cite{chung2016lip,petridis2017end,chung2017lip}, visually indicated sound separation~\cite{ephrat2018looking,gao2018learning,owens2018audio,zhao2018sound,zhao2019sound,xu2019recursive,gan2020music}, audio-visual event localization~\cite{tian2018audio,lin2019dual,wu2019DAM,ramaswamy2020makes,ramaswamy2020see}, audio-visual video parsing~\cite{tian2020unified}, audio-visual embodied navigation~\cite{chen2019soundspaces,gan2020look}, and audio-visual action recognition~\cite{gao2019listentolook,kazakos2019epic,xiao2020audiovisual} to achieve robust auditory or visual perception by integrating audio and visual information. 
However, whether these computational perception models still exhibit robustness under attacks or they are vulnerable to corrupted sensory inputs as in human perception, these have not been systematically evaluated in previous work. 


Inspired by the auditory-visual illusion~\cite{mcgurk1976hearing} in human perception, we present a systematic study on machines' multisensory integration under attacks. We use the audio-visual event recognition task against multimodal adversarial attacks as a proxy to investigate the robustness of audio-visual learning. Adversarial examples are generated with several different attack methods for audio, visual, and both modalities to evaluate the robustness of our models. In addition, different audio-visual fusion methods are explored to validate the correlation between model robustness and multisensory integration. To visually interpret the audio-visual interactions under attacks, we learn a weakly-supervised sound source visual localization model to localize sounding regions in videos.
To mitigate the adversarial multimodal attacks, we propose an audio-visual defense method. It uses external feature memory banks to denoise corrupted features from each modality and learns compact unimodal embeddings by enforcing audio-visual dissimilarity to strengthen invulnerability. For fairly evaluating different defense approaches, we propose a relative improvement (RI) metric that considers results from both clean and attack models and can penalize modality-biased defense models. One audio-visual attack example is illustrated in Fig.~\ref{fig:teaser}.

Extensive experiments can validate that our audio-visual models are susceptible to adversarial perturbations, audio-visual integration could weaken model robustness rather than strengthen under multimodal attacks, even a weakly-supervised sound source visual localization model can be successfully fooled, and the proposed audio-visual defense method can improve network invulnerability without significantly sacrificing clean model performance. 

The main contributions of our work are: (1) systematically investigating the robustness of audio-visual event recognition models against the adversarial multimodal attack with different attackers and fusion methods; (2) qualitatively interpreting the robustness over multimodal attacks in terms of the sound source spatial localization; (3) proposing a novel audio-visual defense method that uses clean external feature memory banks to denoise adversarial audio and visual features and enforces the multimodal dispersion and unimodal embedding compactness to strengthen invulnerability. (4) finding a shortcut of audio-visual defense originating from the modality bias issue and proposing a new evaluation metric: RI.

\section{Related Work}
In this section, we discuss some related work on audio-visual learning, adversarial attack, and adversarial defense. 

\vspace{1mm}
\noindent \textbf{Audio-Visual Learning:}
Audio and visual modalities in videos can provide synchronized and/or complementary information. The multimodal nature of videos enables a series of new and interesting audio-visual learning problems, such as self-supervised audio-visual representation learning~\cite{de1994learning,ngiam2011multimodal,aytar2016soundnet,owens2016ambient,arandjelovic2017look,arandjelovic2018objects,owens2018audio,korbar2018cooperative,hu2019deep}, visually indicated sound separation~\cite{ephrat2018looking,gao2018learning,owens2018audio,zhao2018sound,zhao2019sound,rouditchenko2019self,xu2019recursive,gao2019co,gan2020music,Tian2021cyclic}, vision-infused audio inpainting~\cite{zhou2019vision,morrone2020audio}, sound source spatial localization~\cite{hershey2000audio,kidron2005pixels,senocak2018learning,tian2018audio,arandjelovic2018objects,owens2018audio,qian2020multiple,hu2020discriminative,afouras2020self}, lip reading~\cite{chung2016lip,petridis2017end,chung2017lip}, audio-visual event localization~\cite{tian2018audio,lin2019dual,wu2019DAM,ramaswamy2020makes,ramaswamy2020see}, audio-visual video parsing~\cite{tian2020unified}, audio-visual embodied navigation~\cite{gan2020look,chen2019soundspaces}, audio-visual action recognition~\cite{gao2019listentolook,kazakos2019epic,xiao2020audiovisual,wang2020makes}, and cross-modal generation and prediction~\cite{chung2017you,chen2017deep,zhou2018visual,chen2018lip,chen2019hierarchical,zhou2019talking,gao20192,gao2020visualechoes,zhou2020sep,vasudevan2020semantic,gan2020foley,Zhou2021pose,Xu2021visual}. Although the audio-visual integration with clean data facilitates many audio-visual learning tasks and strengthens model robustness, we do not know whether the robustness still exits when audio and visual modalities are attacked. In this paper, we take audio-visual event recognition as the pretext task to explore audio-visual learning robustness against multimodal adversarial attacks. 

\vspace{1mm}
\noindent \textbf{Adversarial Attack:}
Generating adversarial images to attack deep networks have attracted great interests. A pioneer work is proposed by Szegedy~\etal in~\cite{szegedy2013intriguing}, which uses a box-constrained L-BFGS-based optimization to predict adversarial perturbations for fooling networks. Following the line of the work, many white-box (network architecture and parameters are known) attack approaches are developed to effectively attack image classifiers, including Fast Gradient Sign Method (FGSM)~\cite{goodfellow2014explaining}, iterative FGSM~\cite{kurakin2016adversarial}, DeepFool~\cite{moosavi2016deepfool}, Projected Gradient Descent (PGD)~\cite{madry2017towards}, Jacobian-based Saliency Map Attack (JSMA)~\cite{papernot2016limitations},  Carlini \& Wagner’s attack~\cite{carlini2017towards}, Diverse Input Iterative Attack~\cite{xie2019improving}, and Momentum-based Iterative Method (MIM)~\cite{dong2018boosting}. Building upon research in the visual domain, recent research shows that speech recognition models are also susceptible to adversarial audio examples~\cite{carlini2016hidden,song2017poster,zhang2017dolphinattack,carlini2018audio,qin2019imperceptible,dorr2020towards}. But, how adversarial attacks affect universal sound models has not been answered yet. 

Rather than individual audio and visual adversarial attacks, we investigate audio-visual learning under multimodal attacks, which generate adversarial examples for both audio and visual inputs. Particularly, we explore unconstrained video data from a range of categories (\eg, musical instruments and human activities).

\vspace{1mm}
\noindent \textbf{Adversarial Defense:} 
The adversarial defense aims to improve the invulnerability of deep models under attacks. To counter adversarial attacks, adversarial training approaches~\cite{goodfellow2014explaining,kurakin2016adversarial,tramer2017ensemble,miyato2015distributional} are proposed, which incorporate both clean images and their adversarial counterparts into the training process. Since it is not possible to exploit all different levels of perturbations during adversarial training, the trained models might not be able to generalize to certain unknown attacks. To mitigate adversarial attacks, some approaches~\cite{xie2017mitigating,guo2017countering,sun2019adversarial} apply different pre-processing steps and transformations on the input image. There are also some defense methods that propose new objective functions~\cite{pang2019rethinking,mustafa2019adversarial} to enforce robustness by encouraging compact representations. 
In the audio domain, there are only a few methods~\cite{yang2018characterizing,ma2019detecting} to alleviate adversarial attacks on speech recognition. However, they can only detect adversarial examples and are not able to improve model performance. 

Not competing with state-of-the-art defense methods in the image domain, our goal is to investigate how to take the multimodal nature of audio-visual data into consideration for audio-visual defenses and devise unified defense methods, which can alleviate perturbations from both modalities.


\section{Method}
\label{sec:method}


\subsection{Multimodal Adversarial Attack}
\label{sec:aava}

Let $x_v$ be an input video frame, $x_a$ be an input audio waveform, and $y$ be the corresponding groundtruth label for the multisensory input: $\{x_a, x_v\}$. We denote $\mathcal{F}_{\theta}$ as our audio-visual network, where $\theta$ are the model parameters.

The goal of a multimodal attack is to fool the target multimodal model: $\mathcal{F}_{\theta}$ by adding human imperceptible perturbations into its inputs from multiple modalities, such as audio: $x_a$ and visual: $x_v$ in our problem. Since there are multiple inputs, we can divide our multimodal attack into two categories: \textit{single-modality attacks} that only generate audio adversarial example $x_a^{adv}$ or visual adversarial example $x_v^{adv}$, and \textit{audio-visual attacks} that generate both audio and visual adversarial examples:  $\{x_a^{adv}, x_v^{adv}\}$. 

\noindent \textbf{Adversarial Objective:} To force a trained multimodal model $\mathcal{F}_{\theta}$ to make wrong predictions and the corresponding perturbations be as imperceptible as possible, the objective function for multimodal attacks against
$\mathcal{F}_{\theta}$ with audio and visual inputs is as follows:
\begin{equation}
\begin{aligned}
    \argmax_{x_a^{adv}, x_v^{adv}} &\mathcal{L}(x_a^{adv}, x_v^{adv}, y; \theta)\\ \textrm{s.t.} \quad & ||x_a^{adv} - x_a||_p \leqslant \epsilon_a\\ & ||x_v^{adv} - x_v||_p \leqslant \epsilon_v,
\end{aligned}
\end{equation}
where $\delta_a = x_a^{adv} - x_a$ is the audio adversarial perturbation, $\delta_v = x_v^{adv} - x_v$ is the visual adversarial perturbation, $\mathcal{L}(\cdot)$ is the loss function to optimize $\mathcal{F}_{\theta}$, $||\cdot||_p$ is the $p$-norm, and $\epsilon_a$ and $\epsilon_v$ are audio and visual perturbation budgets, respectively. With the adversarial objective, the attacker will maximize the loss function by seeking small perturbations within allowed budgets, and try to push the trained model to make incorrect predictions. For single-modality attacks, either $\epsilon_a$ or $\epsilon_v$ is 0. In this case, our multimodal model can still access clean inputs from the unattacked modality. For audio-visual attacks, both audio and visual inputs will be corrupted. With exploring effects of different single-modality and audio-visual attacks, we can investigate the model robustness under multimodal attacks. 

\subsection{Audio-Visual Event Recognition}
\label{sec:aver}
We use audio-visual event recognition task as a proxy to explore the audio-visual model robustness under multimodal attacks. Given an audio waveform: $x_a$ and the corresponding video frame: $x_v$ from a short video clip, the goal of the task is to predict the event category of the video clip. To address the problem, we introduce an audio-visual network\footnote{We include details of the architecture in the supplementary material.} as shown in Fig.~\ref{fig:avenet}, which can integrate information from the both modalities to infer event labels.  It uses an 1D convolution-based audio network to extract an audio feature: $f_a \in \mathcal{R}^{d}$ from $x_a$. ResNet~\cite{he2016deep} is adopted as the visual network to extract a visual feature $f_v \in \mathcal{R}^{d}$ from $x_v$. The audio and visual features are integrated by a fusion function outputting a fused feature: $f_{av}$. In practice, we obtain $f_{av} = [f_a; f_v]$ via concatenating the audio and visual features. Taking the $f_{av}$ as an input, a fully-connected layer with a softmax is used to predict its event class probability $p$. The cross-entropy objective function: $ \mathcal{L}_{CE} = -\sum_{i=1}^{k} y_i log(p_i),$
where $k$ is the category number, is used to force the model to learn discriminative features for each class that be mapped to correct output space.

\begin{figure}
    \centering
    \includegraphics[width=\linewidth]{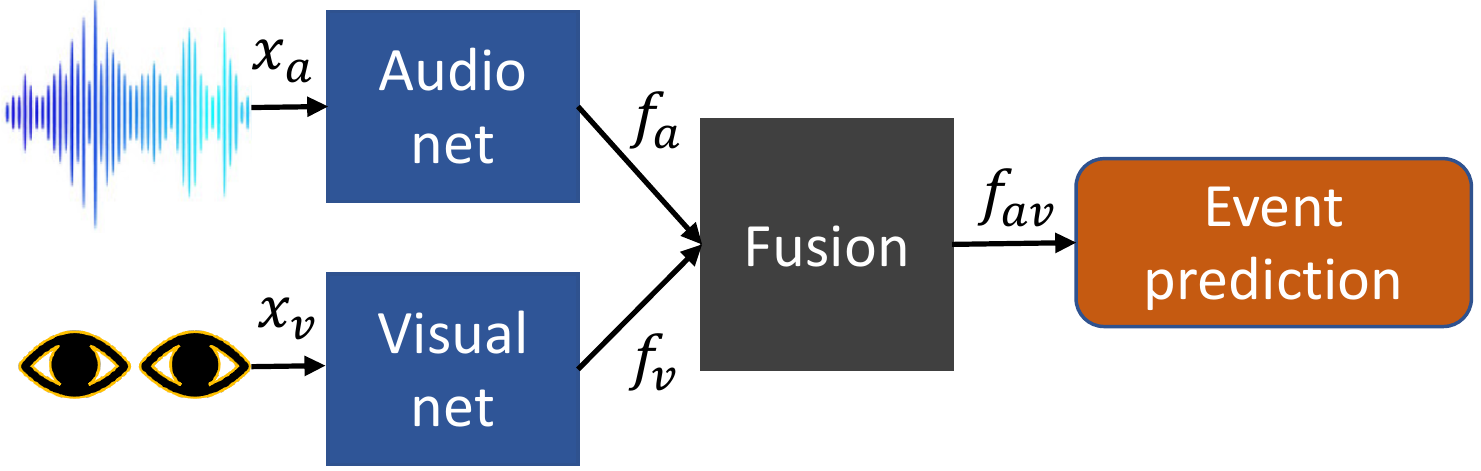}
    \vspace{-5mm}
    \caption{Audio-visual event recognition network. It integrates audio and visual content to predict the event category.} 
    \label{fig:avenet}
    \vspace{-6mm}
\end{figure}

\subsection{Audio-Visual Defense}
\label{sec:avdefense}
 To defend adversaries and improve the robustness of our audio-visual models, we propose an audio-visual defense method. It includes two parts: learning discriminative and compact unimodal embeddings and external feature memory banks for feature denoising. Next, we will describe the details of our audio-visual defense mechanism. 

\subsubsection{Learning Discriminative and Compact Features}
\label{sec:lcave}
Our deep models are threatened by adversarial attacks since the attackers, by maximizing the loss function, will force the output across its originally correct decision region. It has been suggested that high intra-class compactness in the feature space can strengthen the adversarial robustness of classifiers since it makes difficulties for the adversarial attackers to find feasible perturbations within its allowed budget and go beyond the correct decision boundary~\cite{pang2019rethinking,mustafa2019adversarial}. 

Nevertheless, audio and visual data captured by different senses are essentially distinct. The modality gap in our multimodal task makes the encoded audio, and visual features: $f_a$ and $f_v$ from the same input video different, and thus leads intra-class dispersion in the joint audio-visual feature space. Consequently, our audio-visual model becomes susceptible to adversarial perturbations. 
To mitigate the intra-class dispersion and strengthen our model robustness, we should learn more compact audio-visual embeddings. 

 Audio and visual signals that contain synchronized content are ubiquitous, as demonstrated in a wide range of audio-visual tasks~\cite{de1994learning,ngiam2011multimodal,aytar2016soundnet,korbar2018cooperative,tian2018audio,gan2020look}. Motivated from the nature synchronization between the two modalities, it is straightforward to alleviate the intra-class dispersion in the multimodal data by enforcing similarities between audio and visual features. Maximizing the audio-visual similarity can force the model to align the features from the two modalities and project them in a similar feature space, which will decrease the intra-class dispersion accompanying the modality gap reduction. However, the synchronization does not mean that the two modalities are identical. One reason for joint modeling is better than individual modeling is that the additional modalities can provide augmented discriminativeness rather than redundant information. Thus, the similarity constraint might weaken the power of our multimodal models since it decreases discriminative information from individual modalities. To further encourage the multimodal dispersion in the synchronized audio and visual signals, instead of maximizing, we minimize the audio-visual similarity. The objective function is formulated as:
\begin{equation}
    \mathcal{L}_{Sim} = \frac{f_a\cdot f_v}{max(||f_a||_2\cdot||f_v||_2,\eta)},
\end{equation}
where we use the cosine similarity as the measurement and $\eta = 1e-8$ is a small scalar to avoid division by zero. Combining the cross-entropy and similarity losses, we can obtain our final objective function:
\begin{equation}
    \mathcal{L} = \mathcal{L}_{CE} + \mathcal{L}_{Sim}.
\end{equation}
With the $\mathcal{L}_{Sim}$, the model will tend to learn separated audio and visual embeddings. Meanwhile, the $L_{CE}$ will still urge the features to be discriminative, which will implicitly encourage the both separated unimodal embeddings to be more compact and separable. In this manner, we can simultaneously strengthen the multimodal dispersion and embedding compactness to make our audio-visual model more powerful and robust. 

\subsubsection{External Feature Memory Bank}
\label{sec:efm}
When audio and visual inputs are attacked, the features: $f_a^{adv}$ and $f_v^{adv}$ from corresponding audio and visual adversarial examples become noisy and not reliable. To further defend the attackers, we can estimate cleaner audio and visual features: $f_a^{*}$ and $f_v^{*}$ to replace $f_a^{adv}$ and $f_v^{adv}$. 

Inspired by conventional sparse representation-based image restoration approaches~\cite{elad2006image,yang2010image}, we propose to adopt external feature memory banks to denoise attacked audio and visual examples at a feature level. Since audio and visual features are reliable in training data, we use them to build audio and visual external feature memory banks: $M_a\in\mathcal{R}^{d\times K}$ and $M_v\in\mathcal{R}^{d\times K}$, respectively, where $M_a[:, k]$ and $M_v[:, k]$ are audio and visual feature vectors from the same video, and we sample totally $K$ samples. To estimate clean features, the adversarial features are first encoded with the external feature memory banks:
\begin{equation}
    \begin{aligned}
    \min_{\alpha_a} ||f_a^{adv} - M_a\alpha_a||_2^2 + \lambda_a||\alpha_a||_1,\\
    \min_{\alpha_v} ||f_v^{adv} - M_v\alpha_v||_2^2 + \lambda_v||\alpha_v||_1,
    \end{aligned}
\label{eq:lasso}
\end{equation}
where the parameters: $\lambda_a$ and $\lambda_v$ balance sparsity of the solutions and fidelity of the approximation, and $\alpha_a$ and $\alpha_v$ are predicted audio and visual coefficients, respectively. Then, the more reliable audio and visual features can be reconstructed by the corresponding encoded coefficients: $f_a^{*} = M_a\alpha_a$ and $f_v^{*} = M_v\alpha_v$. We solve the Lasso~\cite{tibshirani1996regression} problems in Eq.~\ref{eq:lasso} using the differentiable Iterative Shrinkage Thresholding Algorithm (ISTA)~\cite{gregor2010learning}. 

With the discriminative, compact, and cleaner audio and visual embeddings, our audio-visual model will be more invulnerable to potential multimodal adversarial attacks.


\section{Experiments}

\subsection{Datasets}
\label{subsec:dataset}
\noindent
We use two widely used audio-visual datasets: MIT-MUSIC and Kinetics-Sounds for training and evaluation. 

\noindent
\textbf{MIT-MUSIC:} This dataset~\cite{zhao2018sound} contains clean audio-visual synchronized musical recordings, which covers 11 instrument categories: accordion, acoustic guitar, cello, clarinet, erhu, flute, saxophone, trumpet, tuba,
violin, and xylophone. 520 available videos with solos in the dataset are used to conduct experiments. We randomly divide the data into trian/val/test splits of 312/104/104 videos, respectively. 

\noindent
\textbf{Kinetics-Sounds:} The dataset is a subset of the Kinetics dataset~\cite{carreira2017quo}, which contains YouTube videos with manually annotated human actions. This subset\footnote{Kinetics-Sounds is firstly used in~\cite{arandjelovic2017look}. Since some videos in the subset are not available on the Internet, the downloaded dataset is slightly smaller.} contains 15516 10 second video clips (9309 training, 3104 validation, 3103 test) in 27 human action categories. Rather than only musical instruments, it includes diverse human activities (\eg, chopping wood, ripping paper, tap dancing, and singing). Besides the diversity of scenes, Kinetics-Sounds is more noisy than the MIT-MUSIC, in which audio and visual content inside some videos might not be completely related. 

\begin{figure*}[t]
\captionsetup[subfigure]{labelformat=empty}
\begin{center}
  \begin{subfigure}[b]{0.33\linewidth}
  \includegraphics[width=\linewidth]{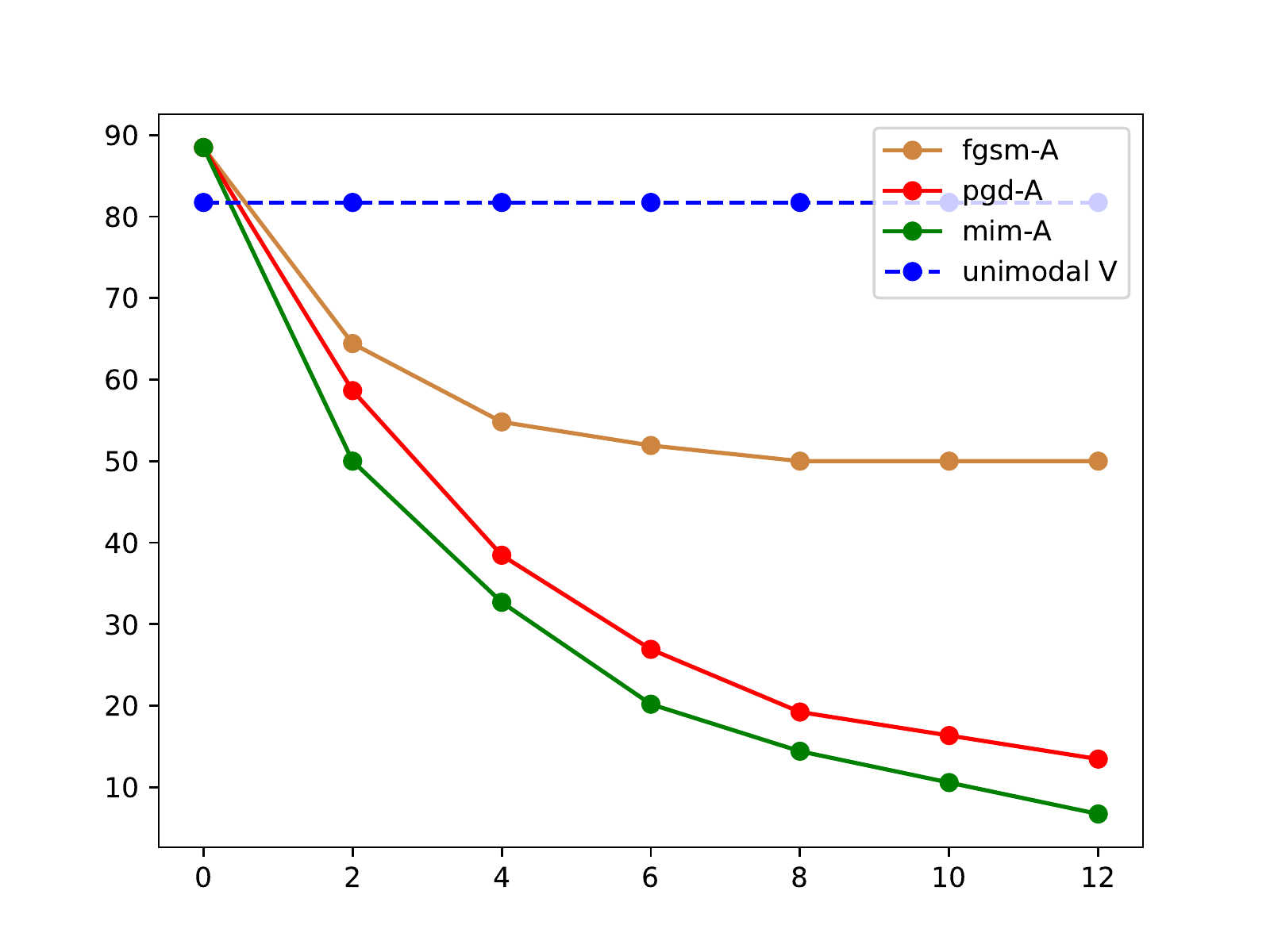}
  \subcaption{(a) Audio Attack}
  \end{subfigure}
  \begin{subfigure}[b]{0.33\linewidth}
  \includegraphics[width=\linewidth]{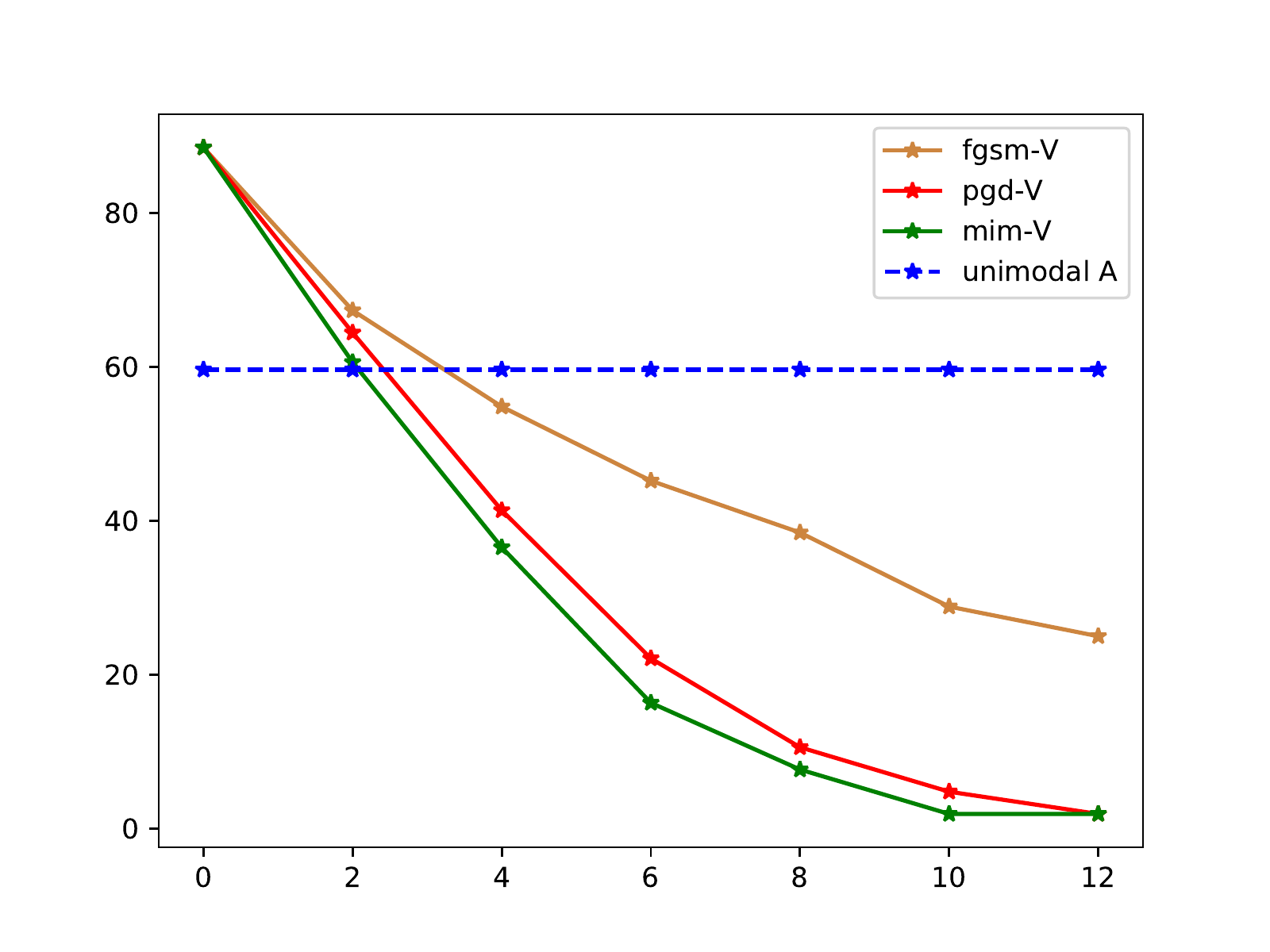}
  \subcaption{(b) Visual Attack}
  \end{subfigure}
  \begin{subfigure}[b]{0.33\linewidth}
  \includegraphics[width=\linewidth]{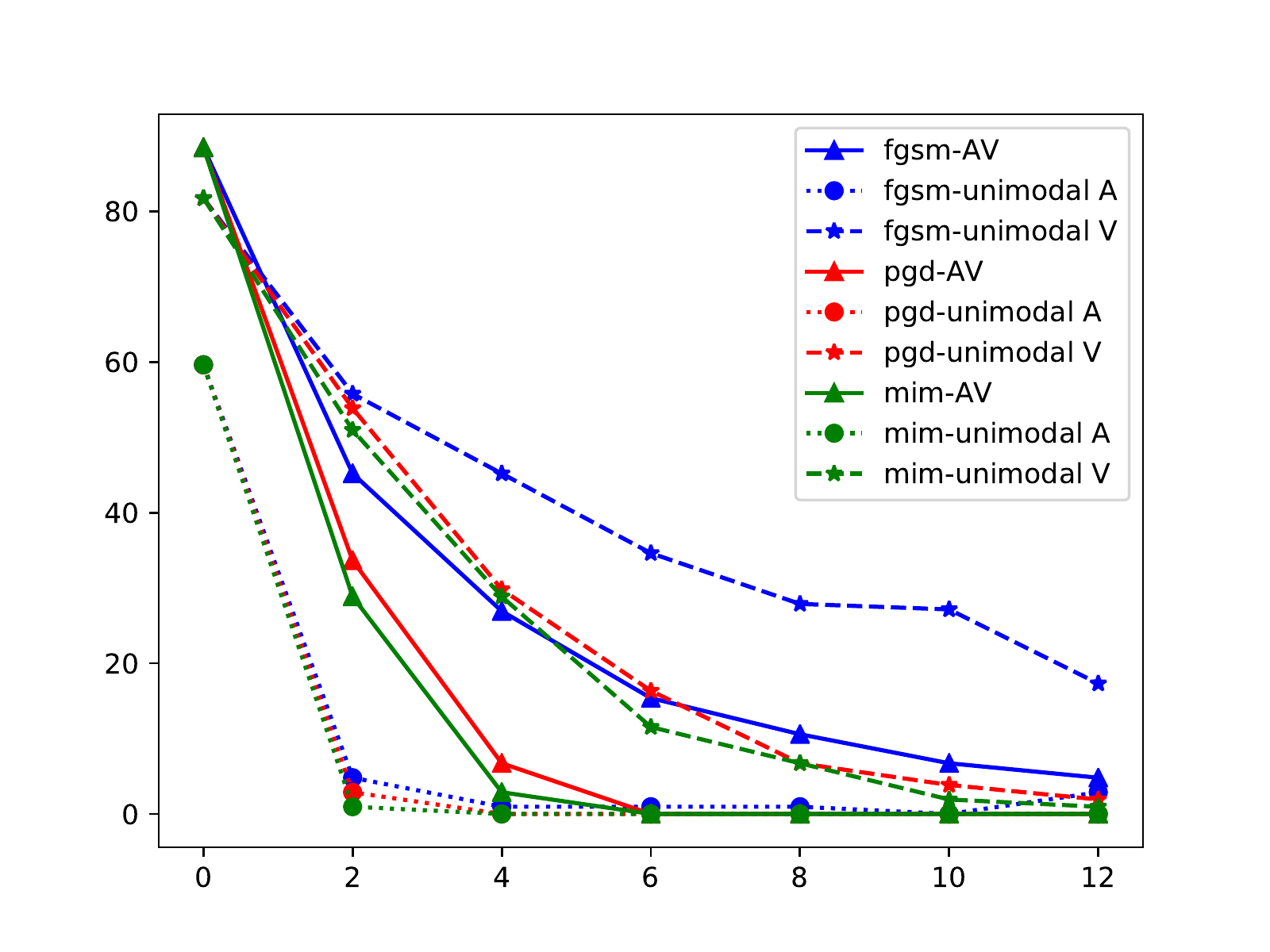}
  \subcaption{(c) Audio-Visual Attack}
  \end{subfigure}
\end{center}
\vspace{-5mm}
   \caption{Adversarial robustness against multimodal attacks on the MIT-MUSIC. The x-axis denotes the attack strength ($\times10^{-3}$) and we set $\epsilon_a = \epsilon_v$ in the audio-visual attack for a better illustration. For the single-modality attack, the attacked audio-visual models in (a) and (b) still have clean visual and audio information, respectively. But, when adversarial perturbations become larger, joint perception models with one attacked modality become even worse than the corresponding individual perception models. Thus, an unreliable modality could weaken perception by the other modality in audio-visual models. A similar observation can also be found in the audio-visual attack (\eg., -AV vs. -unimodal V).
   }
 \label{fig:acc_vs_attack}
\vspace{-3mm}
\end{figure*}

\subsection{Attack Methods}
\label{subsec:attack_method}

\noindent
We evaluate the audio-visual model robustness with $l_{\infty}$-bounded adversarial perturbations, which is widely used as a standard evaluation metric for adversarial robustness~\cite{madry2017towards}. Three different attack methods are used. 

\noindent \textbf{FGSM:} The fast gradient sign method (FGSM)~\cite{goodfellow2014explaining} computes the gradients of the network to generate adversarial examples $x_{{adv}}$ by  $x^{adv} = x + \epsilon\cdot \text{sign}(\nabla_{x}\mathcal{L}(x,y;\theta))$, where $x_{adv}$ is the generated adversarial example, $x$ is the original input, $y$ is the original label, $\theta$ refers to model parameters,  $\epsilon$ is the maximum adversarial perturbation value, and $\mathcal{L}$ is the loss function. For our audio-visual model, we can obtain audio and visual adversarial examples: $x_{adv}^a$ and $x_{adv}^v$ in terms of  $x_{a}^{adv} = x_a + \epsilon_{a}\cdot \text{sign}(\nabla_{x_a}\mathcal{L}(x_a, x_v, y;\theta))$ and  $x_{v}^{adv} = x_v + \epsilon_{v}\cdot \text{sign}(\nabla_{x_v}\mathcal{L}(x_a, x_v, y;\theta))$, respectively.

\noindent \textbf{PGD:} Projected Gradient Descent (PGD)~\cite{madry2017towards} is an iterative variant of the FGSM. We can perform multi-step attacks based on PGD and generate audio and visual adversarial examples with respect to $\epsilon_{a}$ and $\epsilon_{v}$, respectively.

\noindent \textbf{MIM:} Momentum-based Iterative Method (MIM)~\cite{dong2018boosting} integrates a momentum term into the iterative process to further stabilize update directions and mitigate local minima.

\begin{table}[t]
\begin{center}
\scalebox{0.55}{
\begin{tabular}{l| c| c | c c c c|cc}
\toprule
Dataset&Attack &\ding{51}AV &\ding{55}A  &\ding{55}V &  \ding{55}AV &Avg.& Unimodal \ding{51}A & Unimodal \ding{51}V\\
\midrule
\multirow{3}{*}{MM}
    &FGSM~\cite{goodfellow2014explaining}&&50.00&25.00&15.38&30.12\\
&PGD~\cite{madry2017towards}&88.46&13.46&\cellcolor{Gray}\textbf{1.92}&\cellcolor{Gray}\textbf{0.00}&5.09&59.62&81.73\\
&MIM~\cite{dong2018boosting}&&\cellcolor{Gray}\textbf{6.73}&\cellcolor{Gray}\textbf{1.92}&\cellcolor{Gray}\textbf{0.00}&\cellcolor{Gray}\textbf{2.88}\\
\midrule
\multirow{3}{*}{KS}
    &FGSM~\cite{goodfellow2014explaining}&&33.38&15.08&8.18&18.88\\
&PGD~\cite{madry2017towards}&72.42&6.22&1.90&0.77&2.96&35.99&66.08\\
&MIM~\cite{dong2018boosting}&&\cellcolor{Gray}\textbf{3.87}&\cellcolor{Gray}\textbf{1.55}&\cellcolor{Gray}\textbf{0.32}&\cellcolor{Gray}\textbf{1.91}\\
\bottomrule
\end{tabular}
}
\vspace{-1mm}
\caption{Audio-visual event recognition accuracy on MIT-MUSIC and Kinetics-Sounds datasets under different attack methods. \ding{55}A, \ding{55}V, and \ding{55}AV denote that only audio, only visual, and both audio and visual inputs for our audio-visual network are attacked, respectively. We set $\epsilon_a$ and $\epsilon_v$ as 0.12 respectively for \ding{55}A and \ding{55}V, and 0.06 for \ding{55}AV. The symbol: \ding{51} means that inputs are clean. The baselines: Unimodal \ding{51}A and Unimodal \ding{51}V models are two single-modality models.}
\label{tbl:attacks}
\end{center}
\vspace{-6mm}
\end{table}

\subsection{Model Robustness under Multimodal Attacks}
\label{subsec:avsc}
We first investigate the model robustness of audio-visual event recognition under multimodal adversarial attacks. Table~\ref{tbl:attacks} shows audio-visual event recognition accuracy on MIT-MUSIC and Kinetics-Sounds datasets under both single-modality and audio-visual attacks with different attackers. 
To better interpret the multimodal robustness, we also include results from two baselines: Unimodal A and Unimodal V, which are two single-modality models and only use audio and visual modalities, respectively. Clearly, all of the three attack methods: FGSM, PGD, and MIM can significantly decrease recognition results, and the MIM achieves the lowest accuracy under different multimodal attacks. The results show that audio-visual models are susceptible to multimodal adversarial attacks, and the MIM is the most effective attack method among the three attackers.

From Table~\ref{tbl:attacks}, we can also see that our clean audio-visual models (\ding{51}AV) are better than both clean single-modality A (Unimodal \ding{51}A) and V (Unimodal \ding{51}V) models, which can validate that audio-visual integration can strengthen perception robustness and improve audio-visual event recognition performance when input modalities are clean and reliable. But, the conclusion might not hold if the audio-visual model is attacked. Next, we will analyze it based on multimodal attack results. 

\begin{figure*}[t]
\captionsetup[subfigure]{labelformat=empty}
\begin{center}
 \begin{subfigure}[b]{0.16\linewidth}
     \includegraphics[width=\linewidth]{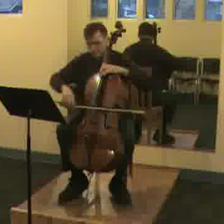}
  \end{subfigure}
  \begin{subfigure}[b]{0.16\linewidth}
  \includegraphics[width=\linewidth]{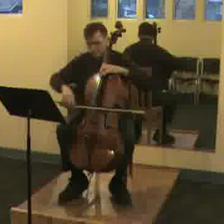}
  \end{subfigure}
  \begin{subfigure}[b]{0.16\linewidth}
     \includegraphics[width=\linewidth]{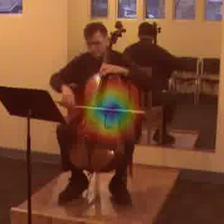}
  \end{subfigure}
  \begin{subfigure}[b]{0.16\linewidth}
  \includegraphics[width=\linewidth]{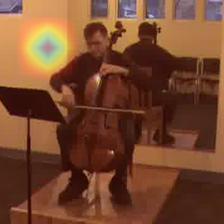}
  \end{subfigure}
  \begin{subfigure}[b]{0.16\linewidth}
  \includegraphics[width=\linewidth]{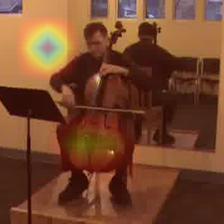}
  \end{subfigure}
  \begin{subfigure}[b]{0.16\linewidth}
  \includegraphics[width=\linewidth]{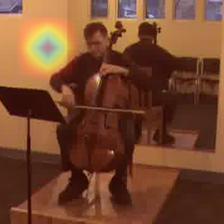}
  \end{subfigure}
  
  
   \begin{subfigure}[b]{0.16\linewidth}
     \includegraphics[width=\linewidth]{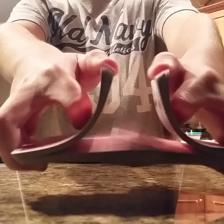}
  \end{subfigure}
  \begin{subfigure}[b]{0.16\linewidth}
  \includegraphics[width=\linewidth]{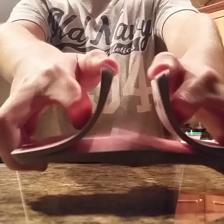}
  \end{subfigure}
  \begin{subfigure}[b]{0.16\linewidth}
     \includegraphics[width=\linewidth]{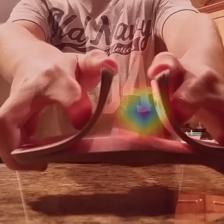}
  \end{subfigure}
  \begin{subfigure}[b]{0.16\linewidth}
  \includegraphics[width=\linewidth]{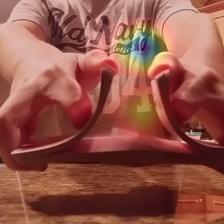}
  \end{subfigure}
  \begin{subfigure}[b]{0.16\linewidth}
  \includegraphics[width=\linewidth]{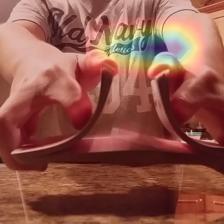}
  \end{subfigure}
  \begin{subfigure}[b]{0.16\linewidth}
  \includegraphics[width=\linewidth]{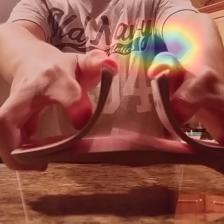}
  \end{subfigure}
  
    \begin{subfigure}[b]{0.16\linewidth}
     \includegraphics[width=\linewidth]{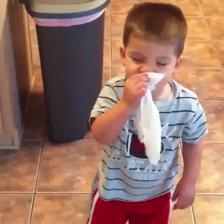}
     \subcaption{Input Frame}
  \end{subfigure}
  \begin{subfigure}[b]{0.16\linewidth}
  \includegraphics[width=\linewidth]{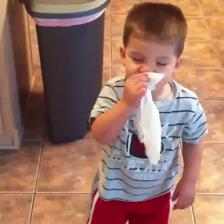}
  \subcaption{Attacked Frame}
  \end{subfigure}
  \begin{subfigure}[b]{0.16\linewidth}
     \includegraphics[width=\linewidth]{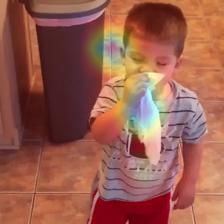}
     \subcaption{w/o Attack}
  \end{subfigure}
  \begin{subfigure}[b]{0.16\linewidth}
  \includegraphics[width=\linewidth]{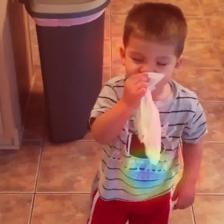}
  \subcaption{Audio Attack}
  \end{subfigure}
  \begin{subfigure}[b]{0.16\linewidth}
  \includegraphics[width=\linewidth]{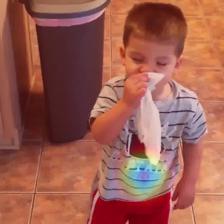}
  \subcaption{Visual Attack}
  \end{subfigure}
  \begin{subfigure}[b]{0.16\linewidth}
  \includegraphics[width=\linewidth]{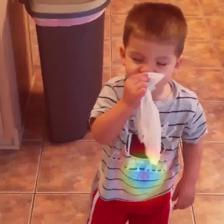}
  \subcaption{AV Attack}
  \end{subfigure}
\end{center}
\vspace{-5mm}
   \caption{Visualizing sound sources under multimodal attacks. The adversarial perturbations in attacked video frames are almost imperceptible. Both single-modality and audio-visual attacks can successfully fool the weakly supervised sound source visual localization model without using sounding object location supervision.
   }
 \label{fig:ssvl_kinetics}
\vspace{-5mm}
\end{figure*}

\noindent \textbf{Single-Modality Attack:} When we use different attackers to perform single-modality attacks on the MIT-MUSIC and Kinetics-Sounds datasets, audio-visual models: \ding{55}A and \ding{55}V are always inferior to Unimodal \ding{51}V and Unimodal \ding{51}A, respectively. For example, the performances drop 91.76\% and 96.77\% on MIT-MUSIC with the MIM attack. 
Note that \ding{55}A and \ding{55}V have clean visual and audio modalities, respectively. The results can demonstrate that audio-visual integration could weaken event recognition performance, when audio or visual inputs are attacked. 

\noindent \textbf{Audio-Visual Attack:} Obviously, when inputs from the both modalities are added adversarial perturbations, the audio-visual models: \ding{55}AV obtain even worse performance than the \ding{55}A and \ding{55}V. When we compare it to attacked unimodal models (see Uimodal A and Unimodal V in Table~\ref{tbl:Defense}), we can see that \ding{55}A of Unimodal A and \ding{55}V of Unimodal V achieve 0.00\% and 11.54\%, while \ding{55}AV of the audio-visual model is 15.38\% under the same FGSM attack on the MIT-MUSIC. Interestingly, the audio-visual model is more invulnerable than the unimodal models against attacks. But when we compare the results from \ding{55}AV of the audio-visual model and \ding{55}V of Unimodal V on the Kinetics-Sounds, joint perception under the audio-visual attack is worse than the visual perception under the single-modality attack.  These results validate that one corrupted modality could still help the other modality, but a joint perception is not always better than individual perceptions under audio-visual attacks. 

Figure~\ref{fig:acc_vs_attack} illustrates the adversarial robustness against multimodal attacks with different perturbations. The results can further validate our findings that audio-visual integration may not always strengthen the audio-visual model robustness under multimodal adversarial attacks. The adversarial robustness of the audio-visual models highly depends on the reliability of the multisensory inputs.

\begin{table}[t]
\begin{center}
\scalebox{0.86}{
\begin{tabular}{ l| c c c c c}
\toprule
Method &\ding{51}AV &\ding{55}A  &\ding{55}V &  \ding{55}AV &Avg.\\
\midrule
Sum&88.46&35.58&45.19&3.85&43.27\\
Concat&88.46&\cellcolor{Gray}\textbf{51.92}&\cellcolor{Gray}\textbf{45.19}&\cellcolor{Gray}\textbf{15.38}&\cellcolor{Gray}\textbf{50.24}\\
FiLM~\cite{perez2017film}&83.65&28.85&39.42&3.85&38.95\\
Gated-Sum~\cite{kiela2018efficient}&\cellcolor{Gray}\textbf{89.42}&33.65&44.23&4.81&43.03\\
Gated-Concat~\cite{kiela2018efficient}&\cellcolor{Gray}\textbf{89.42}&{45.19}&43.27&13.46&47.84\\
\bottomrule
\end{tabular}
}
\vspace{-1mm}
\caption{Audio-visual event recognition accuracy with different fusions on the MIT-MUSIC under FGSM attacks.}
\label{tbl:fusion}
\end{center}
\vspace{-9mm}
\end{table}

\subsection{Audio-Visual Fusions Against Attacks}
\label{subsec:avf}
Audio-visual fusion strategy is important for the performance of our multimodal model. Here, we are curious about whether different audio-visual fusions would also affect the adversarial robustness.  
To answer the question, we compare several different audio-visual fusion approaches: Sum, Concatenation (Concat), FiLM~\cite{perez2017film}, Gated Sum (Gated-Sum)~\cite{kiela2018efficient}, and Gated Concatenation (Gated-Concat)~\cite{kiela2018efficient}, where FiLM and Gated-Sum mix updated audio and visual information together as the Sum before the final prediction layer and the Gated-Concat still preserve the individual information as the Concat.
Table~\ref{tbl:fusion} show audio-visual event recognition results with different fusion methods against FGSM attacks on the MIT-MUSIC dataset. 

From Table~\ref{tbl:fusion}, we can find that our audio-visual models with Sum, Concat, Gated-Sum, and Gated-Concat fusion mechanisms achieve competitive performance on attack-free inputs, and FiLM is worse than the other fusion approaches; audio-visual models: \ding{55}A and \ding{55}V with different fusions achieve inferior performance than Unimodal \ding{51}V and Unimodal \ding{51}A, respectively. The results further support that audio-visual integration could decrease event recognition performance when input audio or visual modalities are not reliable. Another interesting observation is that the Concat and Gated-Concat are much better than the Sum, FiLM, and Gated-Sum under audio-visual attacks, and Concat is the most robust fusion among the compared methods. From the results, we can learn that more audio-visual interactions inside the fusion function might weaken the audio-visual model robustness against the audio-visual attacks.

\subsection{Visualizing Sound Sources Under Attacks}
\label{subsec:ssvl}
To visually interpret the audio-visual interactions under multimodal adversarial attacks, we visualize sound sources in video frames. 
To localize sound sources, we train a weakly-supervised sound source visual localization network. It uses audio-visual event recognition as the pretext task and adopts an audio-guided visual attention mechanism similar to ~\cite{tian2018audio,senocak2018learning} as the localization module. Concretely, we obtain a $N\times N$ visual feature map:$F_v = [f_v^1; ...; f_v^{N^2}] \in \mathcal{R}^{N^2\times d}$ from an input frame: $x_v$ the ResNet~\cite{he2016deep}.  Given the audio feature vector: $f_a$ and $F_v$, we compute audio-guided visual attention weights for each spatial position: $w_i = \frac{exp(f_a^Tf_v^i)}{\sum_j{exp(f_a^Tf_v^j)}}$ and obtain the attended visual feature $f_v^{att} = \sum_i{w_if_v^{i}}$ to replace $f_v$ in the original audio-visual event recognition network. With optimization, the model will force the attention weights to learn to localize sounding visual regions.
Figure~\ref{fig:ssvl_kinetics} illustrates attacked frames and localized sound sources under attacks. 

Without attacks, we can see that our localization model can successfully discover the corresponding sounding regions for different events: \textit{playing cello}, \textit{shuffling cards}, and \textit{blowing noise}. From the generated adversarial frames, we can not find perceptible perturbations. But, the model with the attacked frames fails to localize sound sources. Similarly, the model is fooled by the audio and audio-visual attack. The results demonstrate that weakly-supervised sound source localization models can be attacked even without requiring access to any localization losses for an attacker.

\begin{table}
\begin{center}
\scalebox{0.76}{
\begin{tabular}{ l| c| c c c c|c}
\toprule
Defense (MUSIC) &\ding{51}AV &\ding{55}A  &\ding{55}V &  \ding{55}AV &Avg &RI\\
\midrule
None& 88.46& 51.92& 45.19&15.38&37.50&0.00\\
    Unimodal A& 59.62&0.00&59.62&0.00&19.87&-46.47\\
    Unimodal V& 81.73&\cellcolor{yellow!60}\textbf{81.73}&11.54&11.54&34.94&-9.29\\
    PCL~\cite{mustafa2019adversarial}&83.65&\cellcolor{yellow!60}\textbf{81.73}&37.50&\cellcolor{blue!10}36.54&\cellcolor{blue!10}51.91&9.60\\
    MaxSim&89.42&52.88&45.19&31.73&43.27&6.73\\
MinSim & \cellcolor{yellow!60}\textbf{91.35} &{70.19} &46.15 &\cellcolor{blue!10}{36.54}&{50.96}&\cellcolor{blue!10}16.35\\
ExFMem&{89.42}&53.85&\cellcolor{blue!10}{50.00}&20.19 &41.34&4.80\\
MinSim+ExFMem &\cellcolor{blue!10}{90.38}&\cellcolor{blue!10}{73.08}&\cellcolor{yellow!60}\textbf{53.85}&\cellcolor{yellow!60}\textbf{42.31}&\cellcolor{yellow!60}\textbf{56.41}&\cellcolor{yellow!60}\textbf{20.83}\\
\midrule
Defense (Kinetics) &\ding{51}AV &\ding{55}A  &\ding{55}V &  \ding{55}AV &Avg. &RI\\
\midrule
    None& \cellcolor{blue!10}{72.42}&36.40&26.35&8.09&23.61&0.00\\
    Unimodal A&35.99&1.87&35.99&1.87&13.24&-46.80\\
    Unimodal V&66.08&\cellcolor{yellow!60}\textbf{66.08}&18.72&18.72&34.50&4.55\\
    PCL~\cite{mustafa2019adversarial}&64.50&\cellcolor{blue!10}63.43&29.28&\cellcolor{yellow!60}\textbf{28.67}&\cellcolor{yellow!60}\textbf{40.46}&8.93\\
MaxSim &71.39&34.95&29.57&21.46&28.66&4.02\\
MinSim &70.88&{52.42}&28.12&{21.62}&{34.05}&\cellcolor{blue!10}8.99  \\
ExFMem &\cellcolor{yellow!60}\textbf{72.71}&41.56&\cellcolor{blue!10}{29.93}&10.44&27.31&3.99\\
MinSim+ExFMem  &{71.33}&{55.96}&\cellcolor{yellow!60}\textbf{30.57}&\cellcolor{blue!10}{24.90}&\cellcolor{blue!10}{37.14}&\cellcolor{yellow!60}\textbf{12.44}\\
\bottomrule
\end{tabular}
}
\vspace{-1mm}
\caption{Audio-visual event recognition accuracy on the MIT-MUSIC and Kinetics-Sounds with different defense methods. Here, we use the FGSM ($\epsilon_a$, $\epsilon_v$ = 0.06) to generate audio and visual adversarial examples. Some models (\eg, Unimodal A, Unimodal V, and PCL) highly rely on only one modality, which absolutely makes them more invulnerable to adversarial attacks for another modality. However, they will fail to obtain good performance on clean audio and visual inputs. To better evaluate the robustness of our multisensory defense models, we need to consider model performance on both clean and attacked data and the potential modality bias issue. Top-2 results are highlighted.}
\label{tbl:Defense}
\end{center}
\vspace{-8mm}
\end{table}

\begin{figure*}
    \centering
    \includegraphics[width=0.95\linewidth]{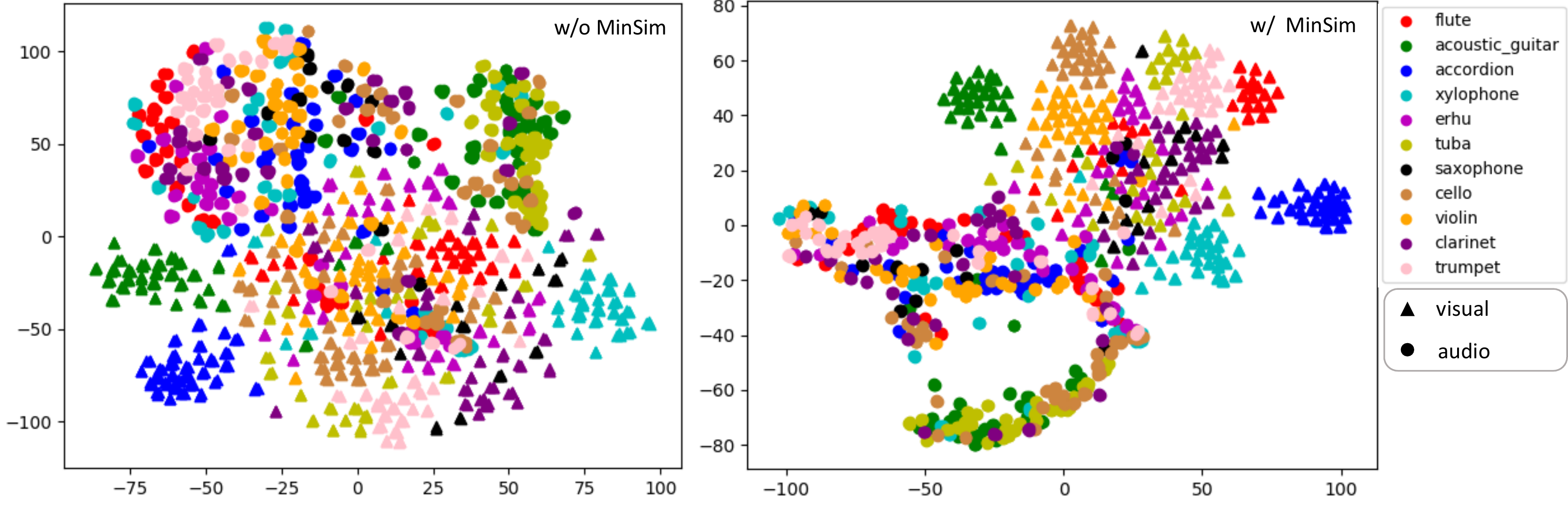}
    \vspace{-1mm}
    \caption{t-SNE visualizations of audio and visual embeddings from w/o MinSim and w/ MinSim models on the MIT-MUSIC. We use symbols: $\blacktriangle$ and $\bullet$ to denote visual and audio modalities, respectively. Different colors refer to different categories. Our MinSim model can learn more intra-class compact and separable embeddings in separated unimodal spaces.} 
    \label{fig:tsne_music}
    \vspace{-1mm}
\end{figure*}

\subsection{Audio-Visual Defense vs. Multimodal Attacks}
\label{subsec:defense}

\noindent \textbf{Baselines:} To validate the effectiveness of the proposed audio-visual defense mechanism, we compare it with several baselines: 1) None: audio-visual network without defense; 2) Unimodal A: audio-only network; 3) Unimodal V: visual-only network; 4) PCL~\cite{mustafa2019adversarial}: a recent state-of-the-art adversarial defense approach, which uses a prototype conformity loss to enforce intra-class compactness and an
inter-class separation; 5) MaxSim: maximizing audio-visual similarity using the $1-\mathcal{L}_{Sim}$ as a loss term to enforce intra-class compactness of joint audio-visual embeddings; 6) MinSim: the proposed dissimilarity constraint to encourage multimodal dispersion and unimodal compactness; 7) ExFMem: the proposed external feature memory banks; 8) MinSim+ExFMem: our full defense model.

\noindent \textbf{Evaluation Metrics:} To evaluate the performance of different defense methods, we use recognition accuracy as the metric. Results from both the clean model: \ding{51}AV and attacked models: \ding{55}A, \ding{55}V, and \ding{55}AV are computed. Since there are multiple defense results under multimodal attacks for a single model, we also use the averaged accuracy: \\
\vspace{-6mm}
\begin{center}
    Avg = $\frac{1}{3}$(\ding{55}A + \ding{55}V + \ding{55}AV),
\end{center}
\vspace{-3mm}
as an overall metric to evaluate different defenses.  However, the metric might not be able to fully reflect the effectiveness of different audio-visual defense methods. For the audio-visual defense, there is a possible shortcut due to the modality bias issue. An audio-visual defense model might mainly make use of information from one dominant  modality. If so, the attacks on another modality will not much affect performance, which might make the defense method achieve pretty good results in terms of the Avg. However, the biased audio-visual defense model fails to joint perception and its \ding{51}AV will achieve worse performance. To address the issue, we propose a relative improvement (RI) metric:
\vspace{-2mm}
\begin{center}
    RI = (\ding{51}$\text{AV}_{m}$ + $\text{Avg}_{m}$) $-$ (\ding{51}$\text{AV}_{n}$ + $\text{Avg}_{n}$),
\end{center}
\vspace{-2mm}
where we consider results from both clean and attacked models, and the $m$ refers to a defense method and $n$ refers to a base model, which is the baseline: None in our experiments. If a defense method decreases clean model performance, the RI will penalize it accordingly.

\noindent \textbf{Results:}  Table~\ref{tbl:Defense} shows defense results of different methods on the MIT-MUSIC and Kinetics-Sound. Although the single-modality model: Unimodal A is not affected by the visual attack, it achieves worse results on the \ding{51}AV and \ding{55}A. We can obtain a similar observation from another modality-biased defense model: Unimodal V. The both defense methods fail to improve robustness on the MIT-MUSIC dataset. 

Interesting results are from the recent defense method: PCL. We can find that the PCL is almost invulnerable to audio attacks (see \ding{51}AV vs. \ding{55}A and \ding{55}V vs. \ding{55}AV) and can also improve the model robustness under visual and audio-visual attacks. From the observation, we can learn that the PCL is a visual-biased defense model. Although the PCL can achieve good results in terms of Avg and even RI, it fails to learn an effective multimodal model. The results further remind us to consider both the modality issue and defense results when we evaluate audio-visual defense methods.

The MaxSim can achieve better performance against audio-visual attacks, however, it is limited in handling single-modality attacks. The results validate that the MaxSim fails to learn compact and powerful unimodal audio and visual embeddings. Compared to the MaxSim, our MinSim is overall more robust against both single-modality and audio-visual attacks. 
Adding the external feature memory bank, the performance of our defense model is further improved. From the results, we can see that our full defense model outperforms all the compared methods on the RI and can achieve comparable or even better clean model performance than the base model.  

To further validate our MinSim defense, we show t-SNE~\cite{maaten2008visualizing} visualizations of learned audio and visual embeddings from w/o MinSim and w/ MinSim in Fig.~\ref{fig:tsne_music}. We can see that our MinSim model learns more intra-class compact and inter-class separable embeddings (especially for the visual) in separated unimodal feature spaces.

\section{Conclusion and Future Work}
In this paper, we investigate the audio-visual model robustness under multimodal attacks. We cast multimodal attacks into two different categories: single-modality attacks and audio-visual attacks. Using the audio-visual event recognition task as a proxy with different fusion and attack methods, we find that audio-visual integration does not always strengthen the perception robustness under multimodal attacks, and it could even decrease performance when the input modalities are not reliable. 

We use the human perception system as a guidance to help us develop computational models. However, there are indeed gaps between AV models and the real perception system and our research is limited by existing learning tools. Humans can perceive events from single modalities when the other modalities are missing. However, our study shows that AV models are susceptible to attacks since they try to exploit information from both modalities fully. Considering the observation and our results, a promising future direction is to design robust AV models that can perform attacked modality-aware predictions.


\noindent
\textbf{Acknowledgement:} We would like to thank the anonymous reviewers for the constructive comments. This work was supported in part by NSF 1741472, 1813709, and 1909912. The article solely reflects the opinions and conclusions of its authors but not the funding agents.

{\small
\bibliographystyle{ieee_fullname}
\bibliography{egbib}
}
\appendix

\definecolor{Gray}{gray}{0.9}
\definecolor{LightCyan}{rgb}{0.88,1,1}

\section*{Appendix}
\noindent 
In this appendix, we first provide more experimental details in Sec.~\ref{sec:exp_detail}. Then, we show more results in Sec.~\ref{sec:exp_result}.

\section{Experimental Details}
\label{sec:exp_detail}
We first introduce an additional dataset: AVE~\cite{tian2018audio} in Sec.~\ref{sec:AVE}. Then, we describe audio and visual data processing in Sec.~\ref{data}. In addition, we give more details of audio and visual networks in Sec.~\ref{arch}. Furthermore, we define the used five different audio-visual fusion functions in Sec.~\ref{sec:fusion}. Finally, more training details are provided in Sec.~\ref{detail}.

\subsection{The AVE Dataset} 
\label{sec:AVE}
Besides the MIT-MUSIC and Kinetics-Sounds datasets, we also explore audio-visual model robustness using another popular audio-visual dataset: AVE~\cite{tian2018audio} to further validate our findings. It consists of 4,143 unconstrained videos spanning 28 event categories. As in \cite{tian2018audio}, we divide the data into train/val/test splits of 3,339/402/402 videos, respectively.

\begin{table*}
\begin{center}
\scalebox{0.83}{
\begin{tabular}{l| c| c | c c c c|cc}
\toprule
Dataset&Attack Methods &\ding{51}AV &\ding{55}A  &\ding{55}V &  \ding{55}AV &Avg.& Unimodal \ding{51}A & Unimodal \ding{51}V\\
\midrule
\multirow{3}{*}{AVE}
    &FGSM~\cite{goodfellow2014explaining}&&40.55&24.88&8.71&24.71\\
&PGD~\cite{madry2017towards}&70.40&20.15&11.44&1.99&11.19&29.85&65.17\\
&MIM~\cite{dong2018boosting}&&15.17&10.20&0.25&8.54\\
\bottomrule
\end{tabular}
}
\vspace{-1mm}
\caption{Audio-visual event recognition accuracy on the AVE dataset under different attack methods ($\epsilon_a$, $\epsilon_v$ = 0.06). \ding{55}A, \ding{55}V, and \ding{55}AV denote that only audio, only visual, and both audio and visual inputs for our audio-visual network are attacked, respectively. The symbol: \ding{51} means that inputs are clean. The baselines: Unimodal \ding{51}A and Unimodal \ding{51}V models are two single-modality models.}
\label{tbl:ave_attack}
\end{center}
\end{table*}
\vspace{2mm}

\subsection{Data Processing}
\label{data}
The sampling rates of sounds and video frames are 11025 Hz and 8 fps, respectively. For each video, we sample a 6$s$ audio clip with 1 video frame at the center position of the sound as the inputs of our audio-visual models. We use a pre-trained ResNet18~\cite{he2016deep} to extract visual features and a 1-D convolution-based model to extract audio features from input audio waveforms. 

\begin{figure}
    \centering
    \includegraphics[width=0.96\linewidth]{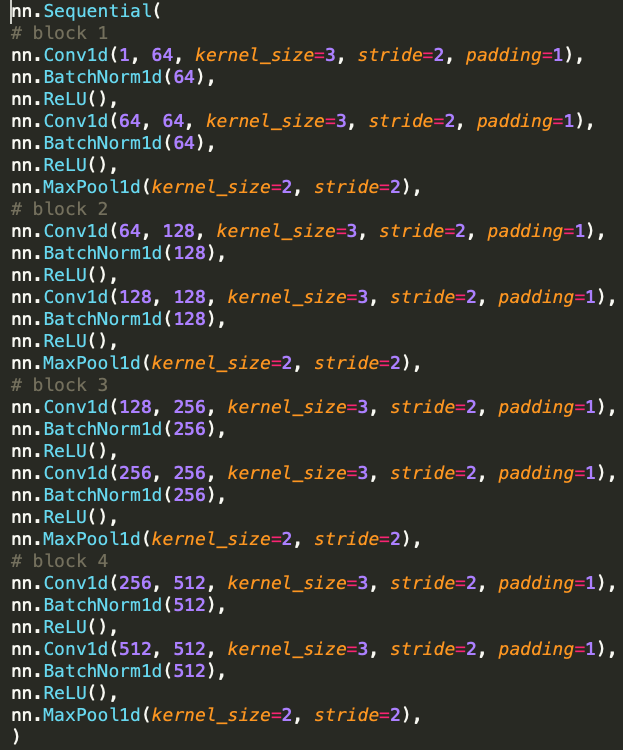}
    \vspace{-1mm}
    \caption{ A Pytorch implemenation of our audio network.} 
    \label{fig:anet}
    \vspace{-1mm}
\end{figure}

\begin{figure*}[t]
\captionsetup[subfigure]{labelformat=empty}
\begin{center}
  \begin{subfigure}[b]{0.33\linewidth}
  \includegraphics[width=\linewidth]{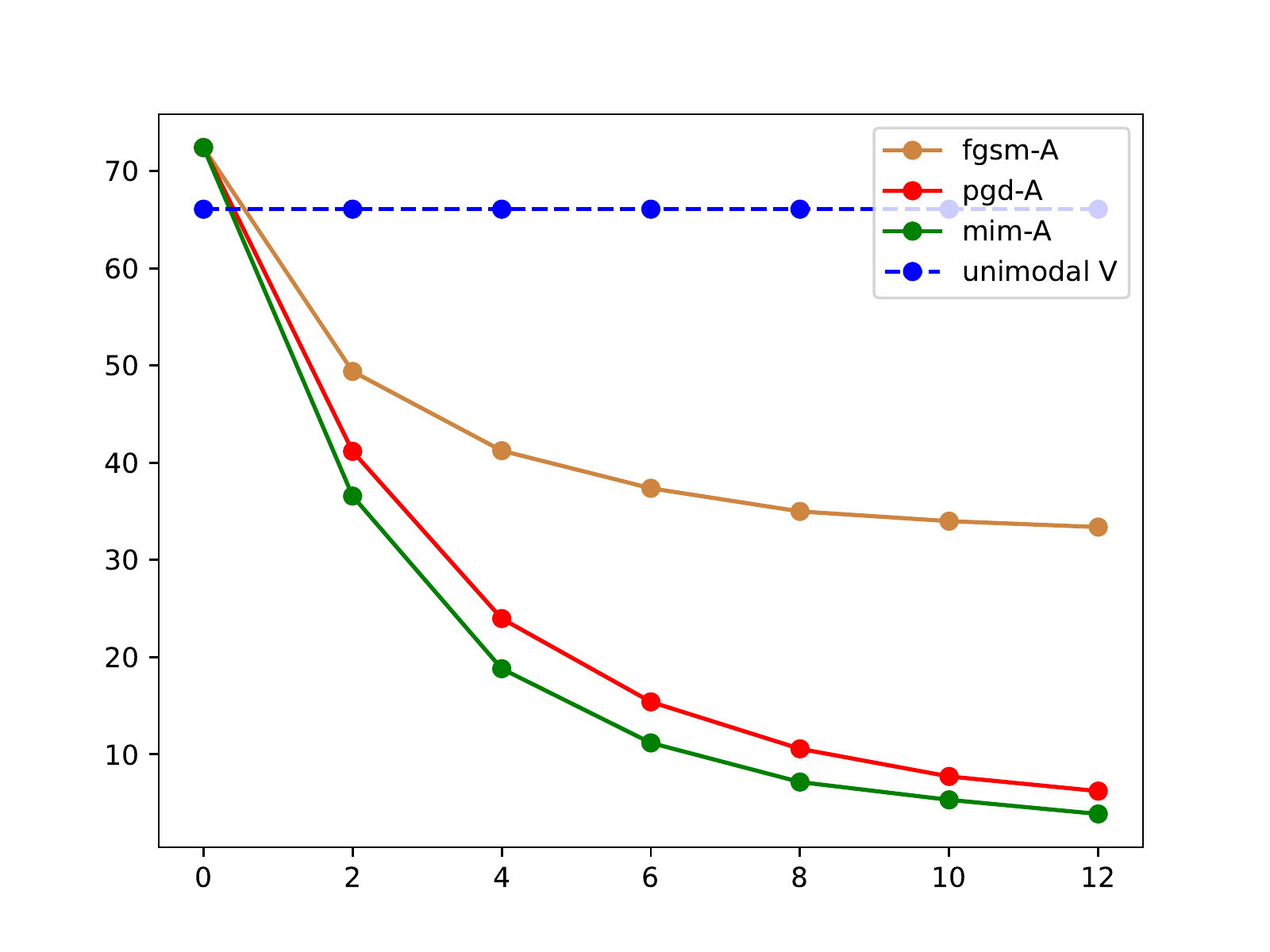}
  \subcaption{(a) Audio Attack}
  \end{subfigure}
  \begin{subfigure}[b]{0.33\linewidth}
  \includegraphics[width=\linewidth]{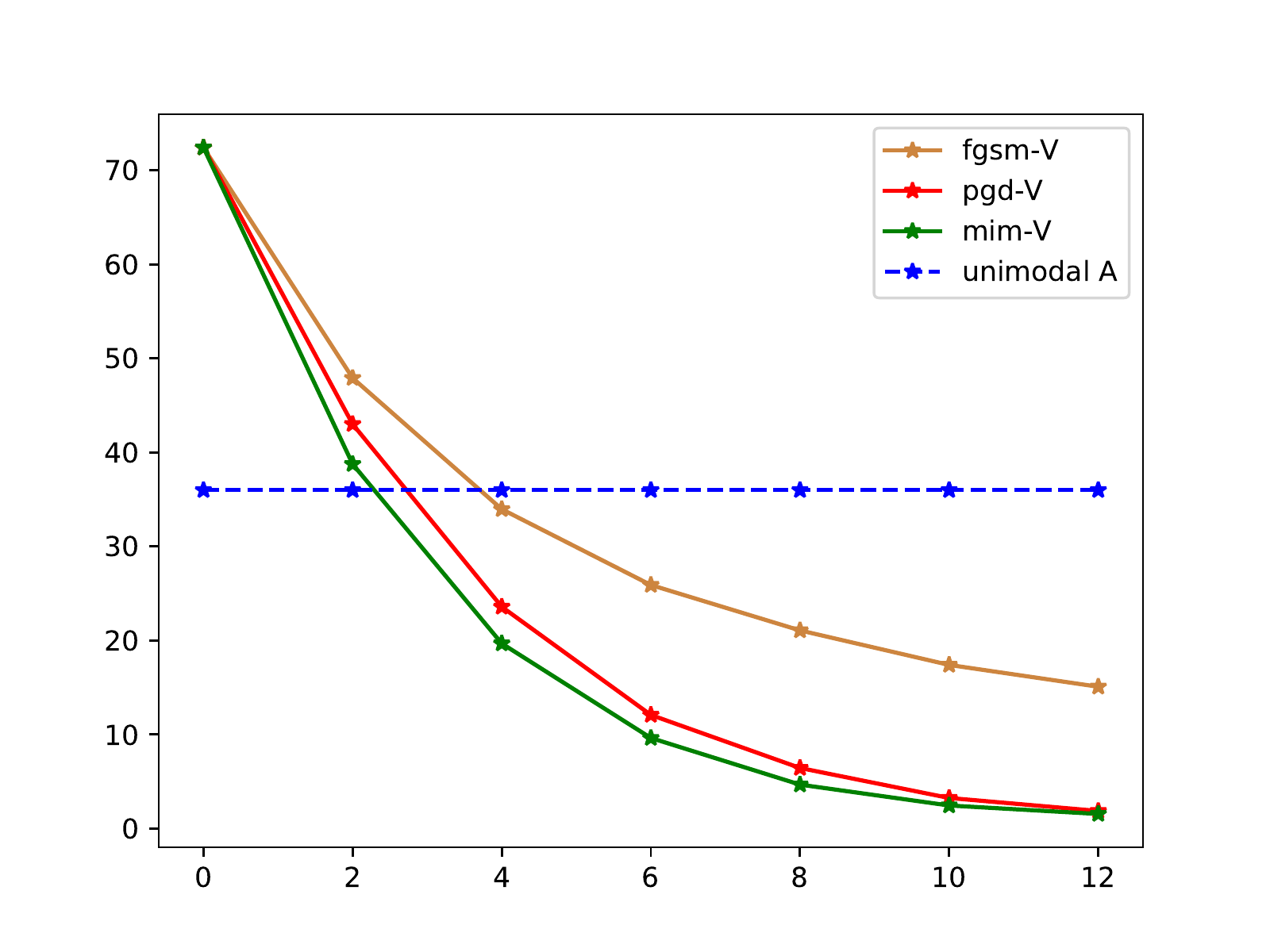}
  \subcaption{(b) Visual Attack}
  \end{subfigure}
  \begin{subfigure}[b]{0.33\linewidth}
  \includegraphics[width=\linewidth]{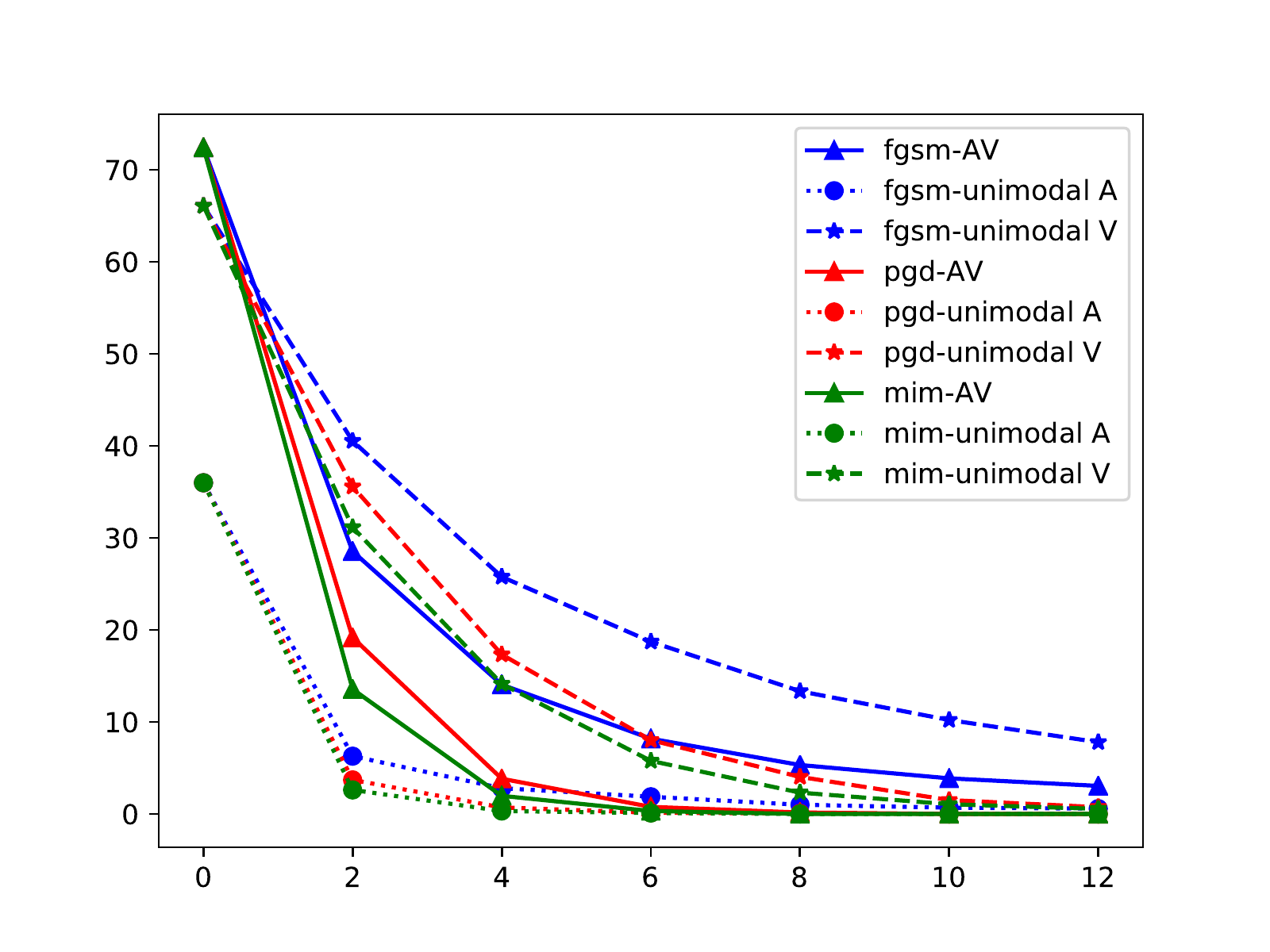}
  \subcaption{(c) Audio-Visual Attack}
  \end{subfigure}
\end{center}
\vspace{-5mm}
   \caption{Adversarial robustness against multimodal attacks on the Kinetics-Sounds. The $x$-axis denotes the attack strength ($\times10^{-3}$) and we set $\epsilon_a = \epsilon_v$ in the audio-visual attack for a better illustration. For the single-modality attack, the attacked audio-visual models in (a) and (b) still have clean visual and audio information, respectively. When adversarial perturbations become larger, joint perception models with one attacked modality become even worse than the corresponding individual perception models. Thus, an unreliable modality could weaken perception by the other modality in audio-visual models. A similar observation can also be found in the audio-visual attack (\eg., -AV vs. -unimodal V). 
   }
 \label{fig:acc_vs_attack_kinetics}
\vspace{-3mm}
\end{figure*}

\begin{figure*}[t]
\captionsetup[subfigure]{labelformat=empty}
\begin{center}
  \begin{subfigure}[b]{0.33\linewidth}
  \includegraphics[width=\linewidth]{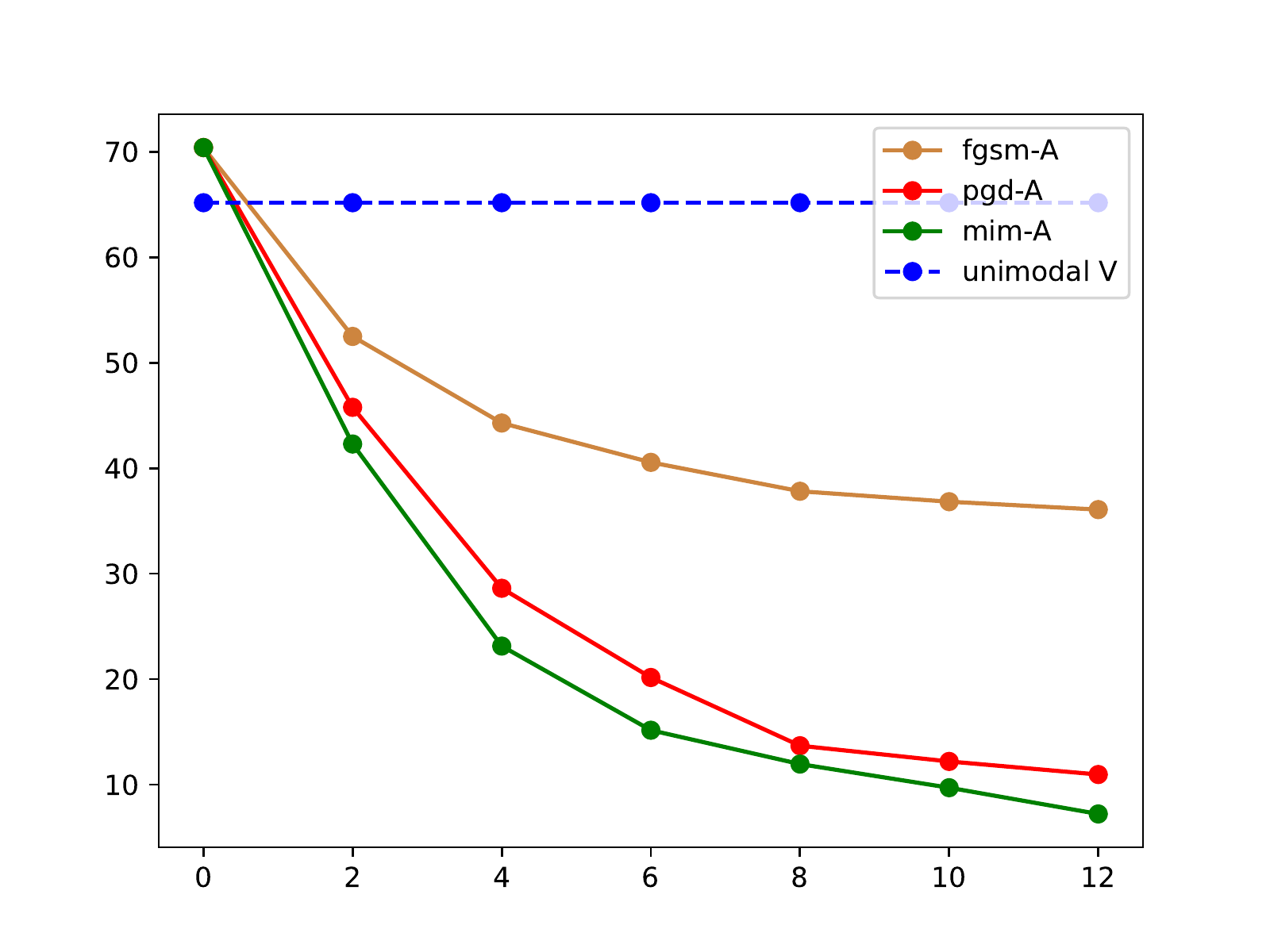}
  \subcaption{(a) Audio Attack}
  \end{subfigure}
  \begin{subfigure}[b]{0.33\linewidth}
  \includegraphics[width=\linewidth]{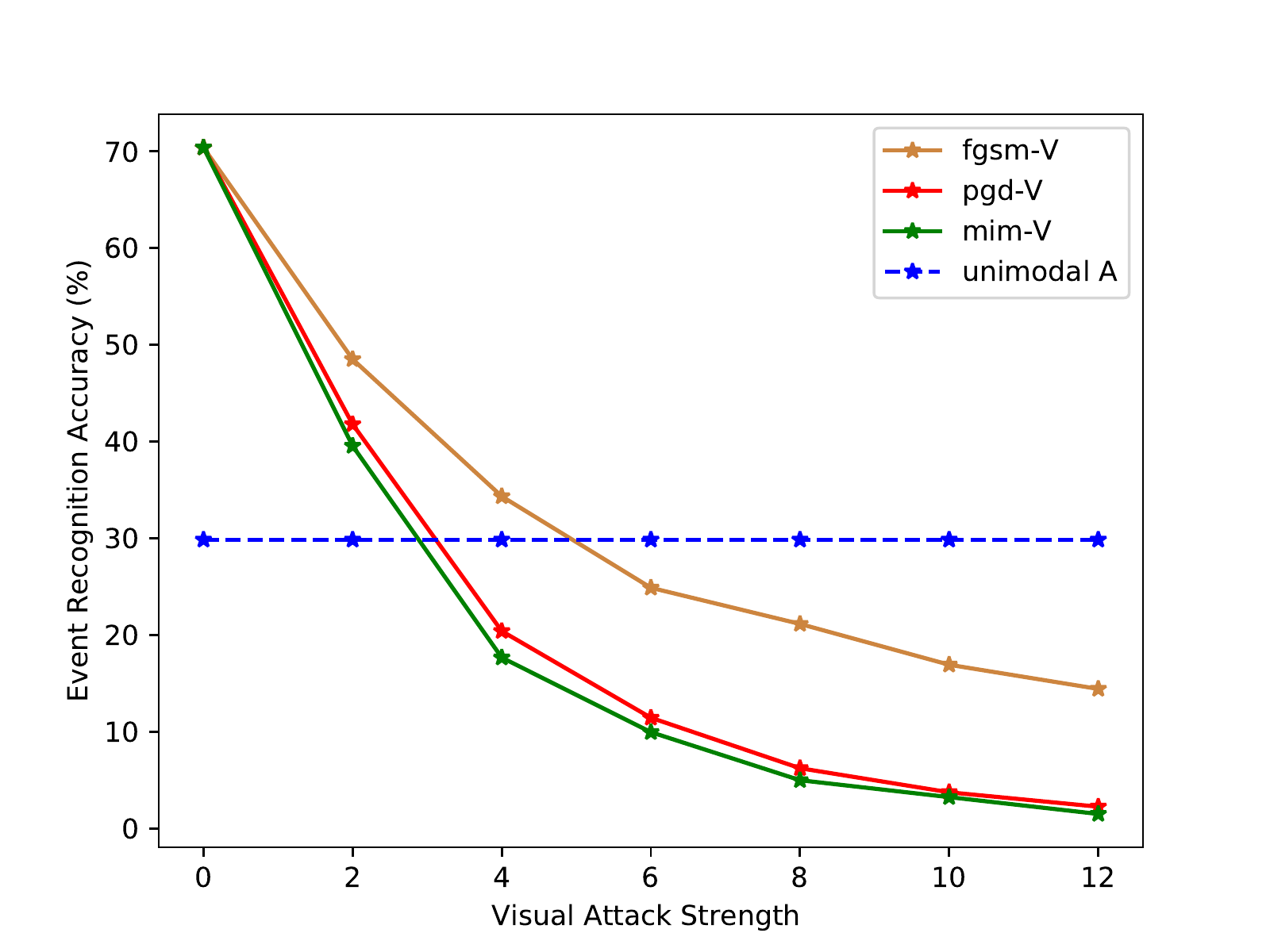}
  \subcaption{(b) Visual Attack}
  \end{subfigure}
  \begin{subfigure}[b]{0.33\linewidth}
  \includegraphics[width=\linewidth]{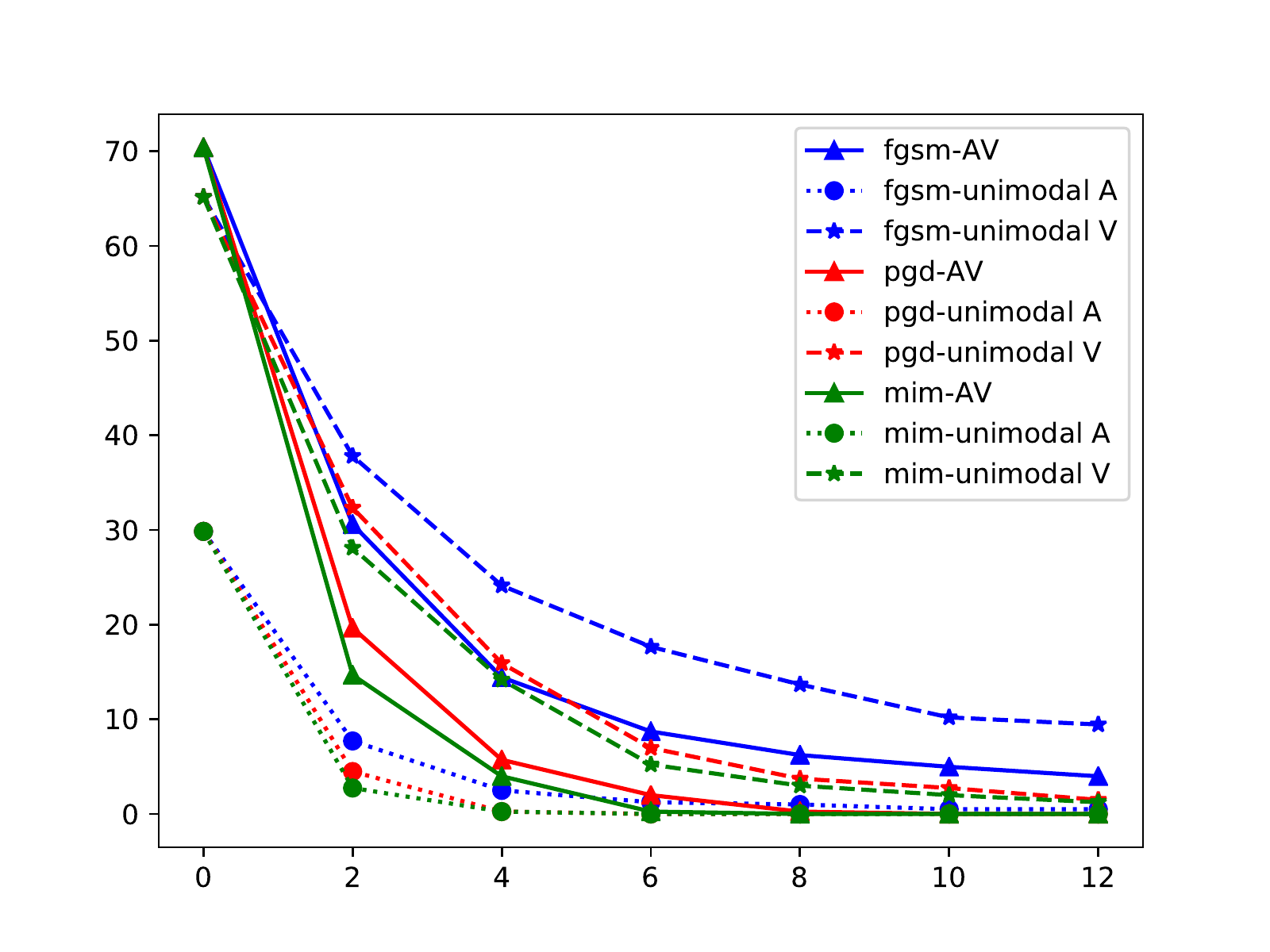}
  \subcaption{(c) Audio-Visual Attack}
  \end{subfigure}
\end{center}
\vspace{-5mm}
   \caption{Adversarial robustness against multimodal attacks on the AVE. The $x$-axis denotes the attack strength ($\times10^{-3}$) and we set $\epsilon_a = \epsilon_v$ in the audio-visual attack for a better illustration. 
   }
 \label{fig:acc_vs_attack_ave}
\vspace{-3mm}
\end{figure*}

\subsection{Architectures}
\label{arch}
\noindent \textbf{Audio Net:} Our audio network takes 6$s$ audio waveforms as inputs and output $512$-D audio feature vectors by a global max pooling after the 1-D Convolution-based network as illustrated in Figure~\ref{fig:anet}. The network consists of 8 convolutional layers in 4 building blocks. 

\vspace{1mm}
\noindent \textbf{Visual Net:} We use the ResNet18~\cite{he2016deep} removing the final Fully-Connected (FC) layer as our visual network. We also obtain 512-D feature vectors by a global max pooling. But, in the weakly-supervised sound source visual localization, we remove the global max pooling to obtain a 2-D feature map for each frame.

\subsection{Audio-Visual Fusion Methods}
\label{sec:fusion}
We use 5 audio-visual fusion methods to explore how different fusion methods affect audio-visual event recognition against multimodal attacks. Here are formulations of the 5 fusion functions. They use an audio feature: $f_a$ and a visual feature: $f_v$ as inputs and obtain a fused feature: $f_{av}$.

\noindent \textbf{Sum:} It directly sums up the features from the both modalities: $f_{av} = f_a + f_v$.

\noindent \textbf{Concat:} The Concat: $f_{av} = [f_a; f_v]$ concatenates the audio and visual features.

\noindent \textbf{FiLM:} The FiLM~\cite{perez2017film} learns to adaptively fuse two different modalities by feature modulations. In our implementation, we use the audio feature as the input of transformation for fusion: $f_{av} = \alpha(f_a)\cdot f_v + \beta(f_a)$, where $\alpha(\cdot)$ is a FC layer and $\beta(\cdot)$ is an identity mapping.

\noindent \textbf{Gated-Sum:} The Gated-Sum~\cite{kiela2018efficient} uses audio and visual features to compute two gates to fuse feature from the other modality, respectively. They can be computed as:
\begin{align}
    f_1 = \sigma(f_a)\cdot f_v,\\
    f_2 = \sigma(f_v)\cdot f_a,
\end{align}
where the $\sigma(\cdot)$ is the Sigmoid function.
The fused features: $f_1$ and $f_2$ are then combined by the Sum: $f_{av} =f _1 + f_2$. 

\noindent \textbf{Gated-Concat:} The Gated-Concat is similar to the the Gated-Sum. It also computes $f_1$ and $f_2$. But, it fuses by a concatenation: $f_{av} = [f_1; f_2]$.

\subsection{Implementation Details}
\label{detail}
We train our network with the standard SGD using 4 NVIDIA 1080TI GPUs.  
 We set the batch size = 48, the initial learning rate of the audio network $= 1e-4$, the initial learning rate of the visual net $= 1e-3$, the initial learning rate of the fusion network with the final FC layer $ = 1e-3$. The epoch numbers are 100, 30, and 100 for the MIT-MUSIC, Kinetics-Sounds, and AVE datasets, respectively. The learning rates drop by multiplying $0.1$ after every 30 epochs for the MIT-MUSIC and AVE and every 10 epochs for the Kinetics-Sounds. For PGD~\cite{madry2017towards} and MIM~\cite{dong2018boosting}, we perform 10-step iterative attacks.
 The parameters: $\lambda_a = \lambda_v = 0.1$. In addition, when we use our external feature memory banks to defend against attacks, we found averaging the denoised and original features can obtain better performance since there are also optimization errors when computing the audio and visual coefficients. 

\begin{figure*}[t]
\captionsetup[subfigure]{labelformat=empty}
\begin{center}
  \begin{subfigure}[b]{0.16\linewidth}
     \includegraphics[width=\linewidth]{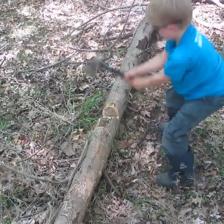}
  \end{subfigure}
  \begin{subfigure}[b]{0.16\linewidth}
  \includegraphics[width=\linewidth]{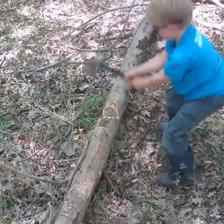}
  \end{subfigure}
  \begin{subfigure}[b]{0.16\linewidth}
     \includegraphics[width=\linewidth]{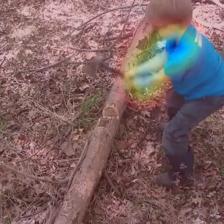}
  \end{subfigure}
  \begin{subfigure}[b]{0.16\linewidth}
  \includegraphics[width=\linewidth]{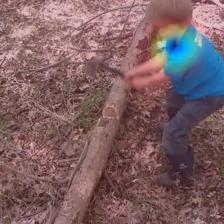}
  \end{subfigure}
  \begin{subfigure}[b]{0.16\linewidth}
  \includegraphics[width=\linewidth]{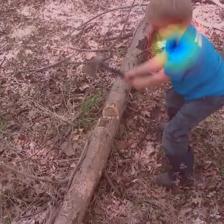}
  \end{subfigure}
  \begin{subfigure}[b]{0.16\linewidth}
  \includegraphics[width=\linewidth]{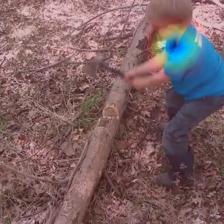}
  \end{subfigure}
  
  \begin{subfigure}[b]{0.16\linewidth}
     \includegraphics[width=\linewidth]{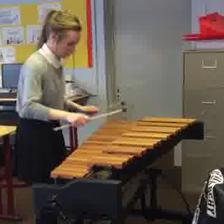}
  \end{subfigure}
  \begin{subfigure}[b]{0.16\linewidth}
  \includegraphics[width=\linewidth]{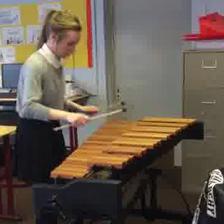}
  \end{subfigure}
  \begin{subfigure}[b]{0.16\linewidth}
     \includegraphics[width=\linewidth]{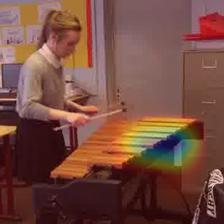}
  \end{subfigure}
  \begin{subfigure}[b]{0.16\linewidth}
  \includegraphics[width=\linewidth]{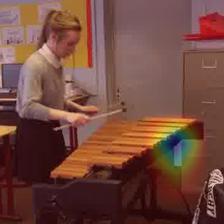}
  \end{subfigure}
  \begin{subfigure}[b]{0.16\linewidth}
  \includegraphics[width=\linewidth]{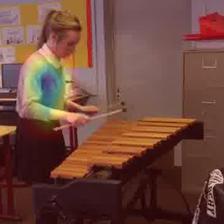}
  \end{subfigure}
  \begin{subfigure}[b]{0.16\linewidth}
  \includegraphics[width=\linewidth]{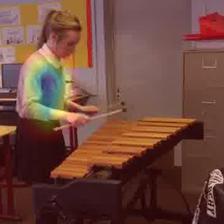}
  \end{subfigure}
  
      \begin{subfigure}[b]{0.16\linewidth}
     \includegraphics[width=\linewidth]{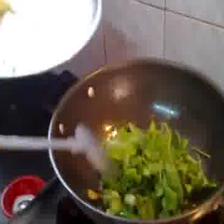}
  \end{subfigure}
  \begin{subfigure}[b]{0.16\linewidth}
  \includegraphics[width=\linewidth]{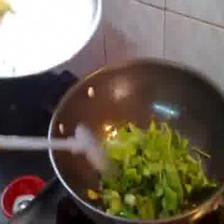}
  \end{subfigure}
  \begin{subfigure}[b]{0.16\linewidth}
     \includegraphics[width=\linewidth]{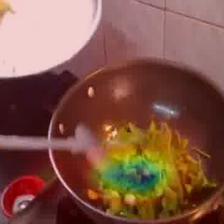}
  \end{subfigure}
  \begin{subfigure}[b]{0.16\linewidth}
  \includegraphics[width=\linewidth]{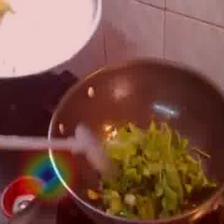}
  \end{subfigure}
  \begin{subfigure}[b]{0.16\linewidth}
  \includegraphics[width=\linewidth]{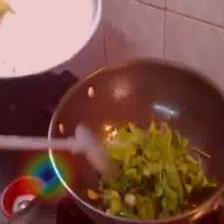}
  \end{subfigure}
  \begin{subfigure}[b]{0.16\linewidth}
  \includegraphics[width=\linewidth]{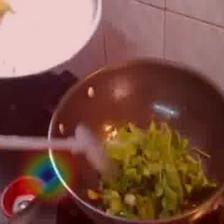}
  \end{subfigure}
  
    \begin{subfigure}[b]{0.16\linewidth}
     \includegraphics[width=\linewidth]{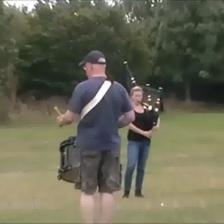}
  \end{subfigure}
  \begin{subfigure}[b]{0.16\linewidth}
  \includegraphics[width=\linewidth]{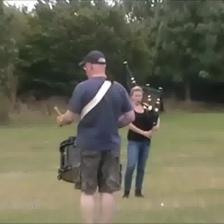}
  \end{subfigure}
  \begin{subfigure}[b]{0.16\linewidth}
     \includegraphics[width=\linewidth]{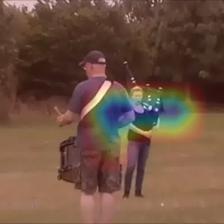}
  \end{subfigure}
  \begin{subfigure}[b]{0.16\linewidth}
  \includegraphics[width=\linewidth]{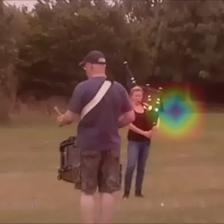}
  \end{subfigure}
  \begin{subfigure}[b]{0.16\linewidth}
  \includegraphics[width=\linewidth]{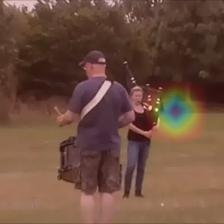}
  \end{subfigure}
  \begin{subfigure}[b]{0.16\linewidth}
  \includegraphics[width=\linewidth]{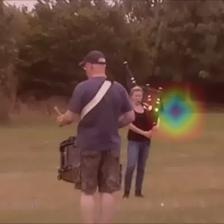}
  \end{subfigure}

      \begin{subfigure}[b]{0.16\linewidth}
     \includegraphics[width=\linewidth]{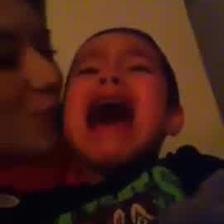}
  \end{subfigure}
  \begin{subfigure}[b]{0.16\linewidth}
  \includegraphics[width=\linewidth]{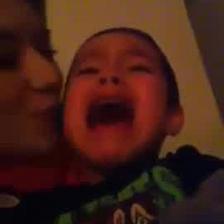}
  \end{subfigure}
  \begin{subfigure}[b]{0.16\linewidth}
     \includegraphics[width=\linewidth]{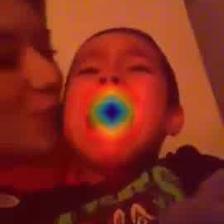}
  \end{subfigure}
  \begin{subfigure}[b]{0.16\linewidth}
  \includegraphics[width=\linewidth]{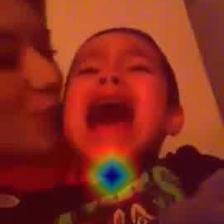}
  \end{subfigure}
  \begin{subfigure}[b]{0.16\linewidth}
  \includegraphics[width=\linewidth]{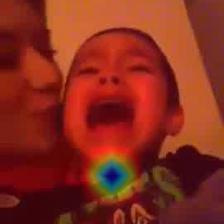}
  \end{subfigure}
  \begin{subfigure}[b]{0.16\linewidth}
  \includegraphics[width=\linewidth]{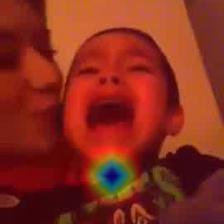}
  \end{subfigure}
  
      \begin{subfigure}[b]{0.16\linewidth}
     \includegraphics[width=\linewidth]{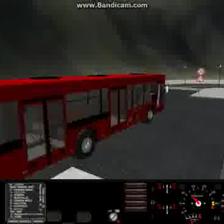}
  \end{subfigure}
  \begin{subfigure}[b]{0.16\linewidth}
  \includegraphics[width=\linewidth]{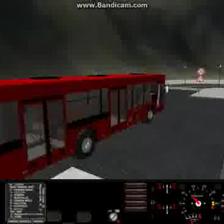}
  \end{subfigure}
  \begin{subfigure}[b]{0.16\linewidth}
     \includegraphics[width=\linewidth]{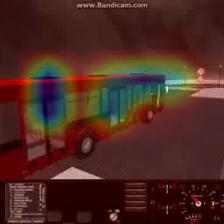}
  \end{subfigure}
  \begin{subfigure}[b]{0.16\linewidth}
  \includegraphics[width=\linewidth]{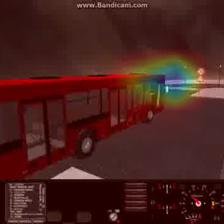}
  \end{subfigure}
  \begin{subfigure}[b]{0.16\linewidth}
  \includegraphics[width=\linewidth]{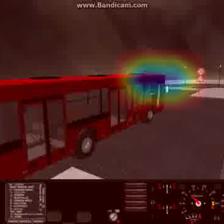}
  \end{subfigure}
  \begin{subfigure}[b]{0.16\linewidth}
  \includegraphics[width=\linewidth]{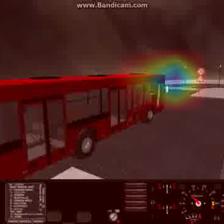}
  \end{subfigure}
  
    \begin{subfigure}[b]{0.16\linewidth}
     \includegraphics[width=\linewidth]{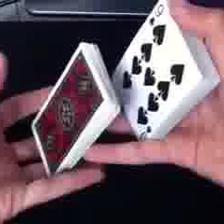}
     \subcaption{Input Frame}
  \end{subfigure}
  \begin{subfigure}[b]{0.16\linewidth}
  \includegraphics[width=\linewidth]{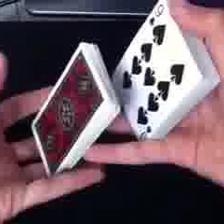}
  \subcaption{Attacked Frame}
  \end{subfigure}
  \begin{subfigure}[b]{0.16\linewidth}
     \includegraphics[width=\linewidth]{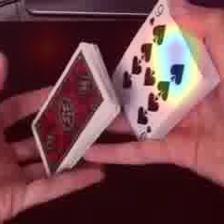}
     \subcaption{w/o Attack}
  \end{subfigure}
  \begin{subfigure}[b]{0.16\linewidth}
  \includegraphics[width=\linewidth]{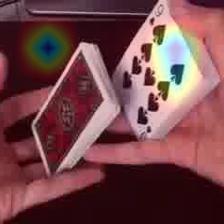}
  \subcaption{Audio Attack}
  \end{subfigure}
  \begin{subfigure}[b]{0.16\linewidth}
  \includegraphics[width=\linewidth]{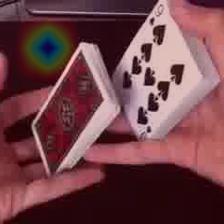}
  \subcaption{Visual Attack}
  \end{subfigure}
  \begin{subfigure}[b]{0.16\linewidth}
  \includegraphics[width=\linewidth]{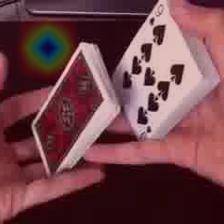}
  \subcaption{AV Attack}
  \end{subfigure}
\end{center}
\vspace{-3mm}
   \caption{Visualizing sound sources under multimodal attacks. The adversarial perturbations in attacked video frames are almost imperceptible. Both single-modality and audio-visual attacks can successfully fool the weakly supervised sound source visual localization model without using sounding object location supervision.
   }
 \label{fig:ssvl}
\end{figure*}

\begin{table}
\begin{center}
\scalebox{0.76}{
\begin{tabular}{ l| c| c c c c|c}
\toprule
Defense (AVE) &\ding{51}AV &\ding{55}A  &\ding{55}V &  \ding{55}AV &Avg &RI\\
\midrule
None& 70.40&40.55&24.88&8.71&24.71&0\\
    Unimodal A& 29.85&1.24&\cellcolor{yellow!60}29.85&1.24&10.77&-54.49\\
    Unimodal V& 65.17&\cellcolor{yellow!60}65.17&17.66&17.66&33.50&3.56\\
    PCL~\cite{mustafa2019adversarial}&61.94&\cellcolor{blue!10}61.69&17.91&17.91&32.50&-0.67\\
    MaxSim&\cellcolor{yellow!60}71.64&35.82&25.62&16.42&25.95&2.48\\
MinSim &70.90&57.21&25.37&\cellcolor{blue!10}21.39&\cellcolor{blue!10}34.66&\cellcolor{blue!10}10.45\\
ExFMem&\cellcolor{blue!10}71.39&44.78&28.11&10.95&27.94&4.22\\
MinSim+ExFMem &\cellcolor{blue!10}71.39&58.21&\cellcolor{blue!10}29.35&\cellcolor{yellow!60}26.62&\cellcolor{yellow!60}38.06&\cellcolor{yellow!60}14.34\\
\bottomrule
\end{tabular}
}
\vspace{-1mm}
\caption{Audio-visual event recognition accuracy on the AVE dataset with different defense methods. Here, we use the FGSM ($\epsilon_a$, $\epsilon_v$ = 0.06) to generate audio and visual adversarial examples. Some models (\eg, Unimodal A, Unimodal V, and PCL) highly rely on only one modality, which absolutely makes them more invulnerable to adversarial attacks for another modality. However, they will fail to obtain good performance on clean audio and visual inputs.  Top-2 results are highlighted.}
\label{tbl:Defense_AVE}
\end{center}
\vspace{-5mm}
\end{table}

\section{Experimental Results}
\label{sec:exp_result}
To further validate our findings, we show more experimental results on audio-visual model robustness under multimodal attacks in Sec.~\ref{robust_ma},  sound source localization under attacks in Sec.~\ref{ssvl}, and audio-visual defense in Sec.~\ref{avd}. 

\subsection{Robustness under Multimodal Attacks}
\label{robust_ma}

We first show the audio-visual event recognition results on the AVE dataset with different attack methods in the Table~\ref{tbl:ave_attack}. Similar to observations on the MIT-MUSIC and Kinetics-Sounds, our audio-visual model can be easily fooled, and the joint perception models: \ding{55}A (with clean visual) and \ding{55}V (with clean audio) are worse than Unimodal \ding{51}V and \ding{51}A, respectively. Thus, audio-visual integration could even weaken event recognition when audio or visual inputs are attacked. We further illustrate results of adversarial robustness against multimodal attacks with different attack strengths on the Kinetics-Sounds and AVE in Figure~\ref{fig:acc_vs_attack_kinetics} and Figure~\ref{fig:acc_vs_attack_ave}, respectively. The results can further validate our findings that audio-visual integration may not always strengthen the audio-visual model robustness under multimodal attacks and the adversarial robustness of the audio-visual models highly depends on the reliability of the multisensory inputs.

\subsection{Sound Source Localization under Attacks}
\label{ssvl}

We show more sound source localization results under multimodal attacks in Figure~\ref{fig:ssvl}. A large range of events (\eg, chopping wood, playing xylophone, frying, baby crying, running bus) are covered. We can see that the weakly-supervised sound source visual localization model is susceptible to both single-modality and audio-visual attacks.

\begin{table}
\begin{center}
\scalebox{0.78}{
\begin{tabular}{ l| c| c c c c|c}
\toprule
Defense (MUSIC) &\ding{51}AV &\ding{55}A  &\ding{55}V &  \ding{55}AV &Avg &RI\\
\midrule
None& 88.46&20.19&16.35&0.00&12.18&0.00\\
    Unimodal A& 59.62&0.00&\cellcolor{yellow!60}59.62&0.00&19.87&-21.15\\
    Unimodal V&81.73&\cellcolor{yellow!60}81.73&11.54&11.54&34.94&16.03\\
    PCL~\cite{mustafa2019adversarial}&83.65&\cellcolor{blue!10}79.81&\cellcolor{blue!10}17.31&17.31&\cellcolor{yellow!60}38.14&\cellcolor{blue!10}21.15\\
    MaxSim&\cellcolor{blue!10}89.42&25.00&31.73&15.38&24.04&12.84\\
    Ours (Full)&\cellcolor{yellow!60}90.38&64.42&\cellcolor{blue!10}27.88&\cellcolor{yellow!60}18.27&\cellcolor{blue!10}36.86&\cellcolor{yellow!60}26.60\\
\toprule
Defense (Kinetics) &\ding{51}AV &\ding{55}A  &\ding{55}V &  \ding{55}AV &Avg &RI\\
\midrule
None& \cellcolor{yellow!60}72.42&11.18&9.63&0.32&7.04&0.00\\
    Unimodal A& 35.99&0.10&35.99&0.10&12.06&-31.41\\
    Unimodal V&66.08&\cellcolor{yellow!60}66.08&5.77&5.77&\cellcolor{blue!10}25.87&12.49\\
    PCL~\cite{mustafa2019adversarial}&64.50&\cellcolor{blue!10}62.98&\cellcolor{yellow!60}20.26&\cellcolor{yellow!60}19.97&\cellcolor{yellow!60}34.40&\cellcolor{yellow!60}19.44\\
    MaxSim&\cellcolor{blue!10}71.39&17.65&\cellcolor{blue!10}18.78&\cellcolor{blue!10}13.89&16.77&8.70\\
    Ours (Full)&71.33&44.72&16.04&9.99&23.58&\cellcolor{blue!10}15.45\\
\toprule
Defense (AVE) &\ding{51}AV &\ding{55}A  &\ding{55}V &  \ding{55}AV &Avg &RI\\
\midrule
None& 70.40&15.17&10.20&0.25&8.54&0.00\\
    Unimodal A& 29.85&0.00&\cellcolor{yellow!60}29.85&0.00&9.95&-39.14\\
    Unimodal V&65.17&\cellcolor{yellow!60}65.17&5.22&5.22&\cellcolor{blue!10}25.20&\cellcolor{blue!10}11.43\\
    PCL~\cite{mustafa2019adversarial}&61.94&\cellcolor{blue!10}61.44&7.21&6.97&25.21&8.21\\
    MaxSim&\cellcolor{yellow!60}71.64&15.17&12.94&\cellcolor{blue!10}8.96&12.36&5.06\\
    Ours (Full)&\cellcolor{blue!10}71.39&52.99&\cellcolor{blue!10}14.43&\cellcolor{yellow!60}11.44&\cellcolor{yellow!60}26.29&\cellcolor{yellow!60}18.74\\
\bottomrule
\end{tabular}
}
\vspace{-1mm}
\caption{Audio-visual defense against the MIM~\cite{dong2018boosting} attack on the MIT-MUSIC, Kinetics-Sounds, AVE datasets. Here, we use the MIM with $\epsilon_a$, $\epsilon_v$ = 0.06 to generate audio and visual adversarial examples. Our full defense method combines the MinSim and ExFMem. Our audio-visual defense method can successfully defend against strong MIM attacks without the modality bias problem.  Top-2 results are highlighted. }
\label{tbl:Defense_mim}
\end{center}
\vspace{-5mm}
\end{table}

\begin{figure*}
    \centering
    \includegraphics[width=0.96\linewidth]{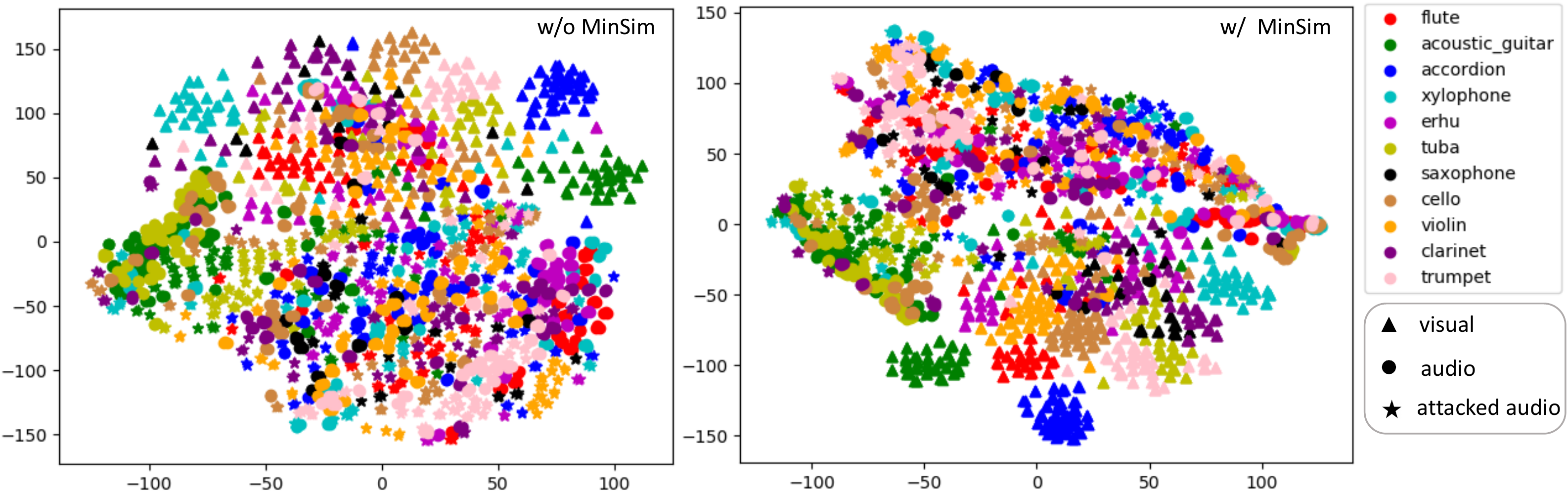}
    \vspace{-1mm}
    \caption{ t-SNE visualizations of audio (clean and attacked) and visual embeddings from w/o MinSim and w/ MinSim models on the MIT-MUSIC. We use symbols: $\blacktriangle$, $\bullet$, and $\star$ to denote visual, audio, attacked audio modalities, respectively. Different colors refer to different categories. Our MinSim model can learn more intra-class compact and separable embeddings in separated unimodal spaces. Thus, the attacked audio samples generated by w/ MinSim are much closer to clean samples in the same categories (\eg, violin, tuba, flute) than the adversarial audio examples obtained by the w/o MinSim.} 
    \label{fig:attacked_a_tsne}
    \vspace{-1mm}
\end{figure*}

\begin{figure*}
    \centering
    \includegraphics[width=0.96\linewidth]{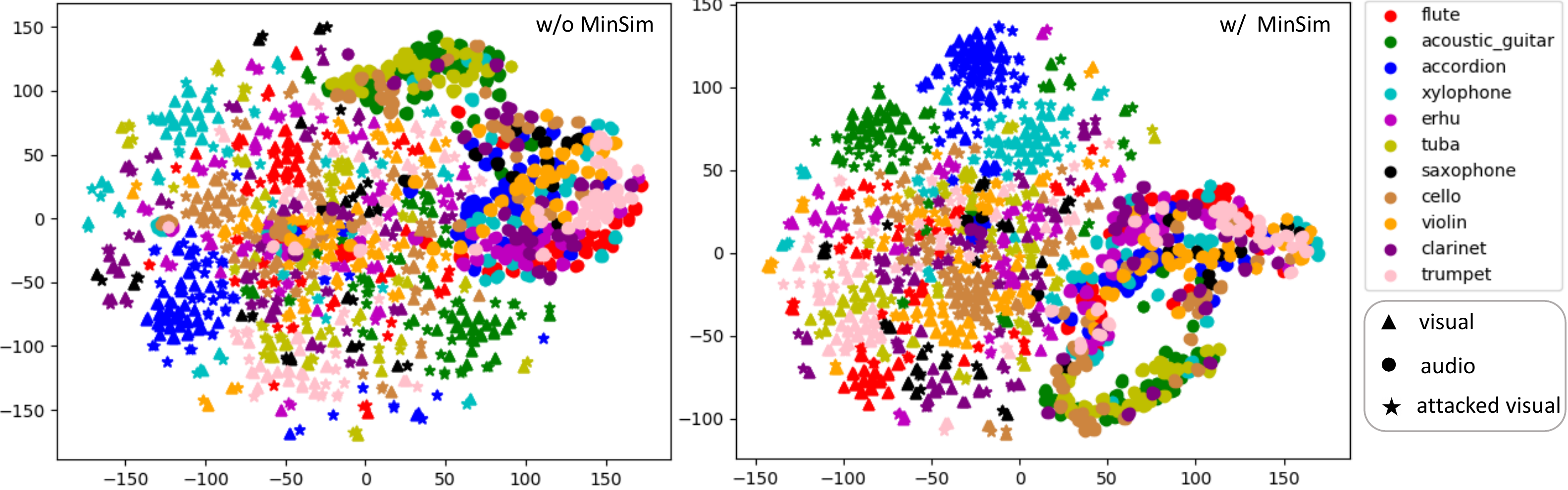}
    \vspace{-1mm}
    \caption{ t-SNE visualizations of audio and visual (clean and attacked) embeddings from w/o MinSim and w/ MinSim models on the MIT-MUSIC. We use symbols: $\blacktriangle$, $\bullet$, and $\star$ to denote visual, audio, attacked visual modalities, respectively. Different colors refer to different categories. The attacked visual samples generated by w/ MinSim are much closer to clean samples in the same categories (\eg, accordion, xylophone, and flute) than the adversarial visual examples obtained by the w/o MinSim.} 
    \label{fig:attacked_v_tsne}
    \vspace{-1mm}
\end{figure*}

\subsection{Audio-Visual Defense}
\label{avd}

 We show defense results against the FGSM attack on the AVE dataset in Table~\ref{tbl:Defense_AVE}. Similar to results on the Kinetics-Sounds and MIT-MUSIC datasets, the proposed: MinSim and ExFMem can improve audio-visual model robustness against both single-modality and audio-visual attacks and our full model outperforms the compared baselines without the modality bias issue.
 
 Since the MIM is the strongest attacker among the three methods, we provide audio-visual defense results against the MIM attacker on the three different datasets in Table~\ref{tbl:Defense_mim} to further demonstrate the effectiveness of our audio-visual defense method. We can see that our method can still
improve audio-visual model robustness against the powerful MIM attacker, and it outperforms all of the compared approaches in terms of the RI on the MIT-MUSIC and AVE. The results further demonstrate that our defense method can generalize to different datasets and defend against different attackers. Moreover, we can find that the two models: Unimodal V and PCL, achieve lower performance on the Kinetics-Sounds for clean audio and visual inputs due to the modality bias problem, while they achieve ``good'' defense results against attacks by the shortcut. The results suggest us to further punish the biased audio-visual models when we evaluate audio-visual defense methods. 

We show t-SNE visualizations of both attacked audio and attacked visual embeddings from w/o MinSim and w/ MinSim in Figure \ref{fig:attacked_a_tsne} and Figure~\ref{fig:attacked_v_tsne}, respectively. We can see that the attacked samples generated by w/ MinSim are closer to clean samples in the same categories than the attacked sampled produced by w/o MinSim, especially for the attacked audio embedding in Figure \ref{fig:attacked_a_tsne}, since the w/ MinSim
can force our audio-visual models to strengthen multimodal dispersion and unimodal compactness. The results can further validate the effectiveness of the proposed MinSim defense mechanism.

\end{document}